\documentclass{article}

\usepackage[final,nonatbib]{neurips_2022}

\usepackage[utf8]{inputenc} % allow utf-8 input
\usepackage[T1]{fontenc}    % use 8-bit T1 fonts
\usepackage{hyperref}       % hyperlinks
\usepackage{url}            % simple URL typesetting
\usepackage{booktabs}       % professional-quality tables
\usepackage{amsfonts}       % blackboard math symbols
\usepackage{nicefrac}       % compact symbols for 1/2, etc.
\usepackage{microtype}      % microtypography
\usepackage{xcolor}         % colors
\usepackage{amsmath}
\usepackage{amsfonts}
\usepackage{amssymb}
\usepackage{xspace}
\usepackage{graphicx}
\usepackage{booktabs}
\usepackage{subcaption}
\usepackage{makecell}
\newcommand{\proj}{LE-PDE\xspace}

\def\u{\mathbf{u}}
\def\x{\mathbf{x}}
\def\X{\mathbb{X}}
\def\R{\mathbb{R}}
\def\U{\mathbb{U}}
\def\a{\mathbf{a}}
\def\z{\mathbf{z}}
\newcommand{\xhdr}[1]{{\noindent\bfseries #1}.}
\usepackage[makeroom]{cancel}

\usepackage{wrapfig}

\newcommand{\projnolatent}{LE-PDE-\cancel{latent}\xspace}

\title{Learning to Accelerate Partial Differential Equations via Latent Global Evolution}

\author{%
   Tailin Wu\\
   Department of Computer Science\\
   Stanford University\\
   \texttt{tailin@cs.stanford.edu} \\
    \And
    Takashi Maruyama\\
    NEC Corp. \& Stanford University\\
    \texttt{49takashi@nec.com \&}\\ \texttt{takashi279@cs.stanford.edu} \\
    \And
    Jure Leskovec\\
    Department of Computer Science\\
   Stanford University\\
  \texttt{jure@cs.stanford.edu} \\
}

\begin{document}

\maketitle

\begin{abstract}

Simulating the time evolution of Partial Differential Equations (PDEs) of large-scale systems is crucial in many scientific and engineering domains such as fluid dynamics, weather forecasting and their inverse optimization problems. However, both classical solvers and recent deep learning-based surrogate models are typically extremely computationally intensive, because of their local evolution: they need to update the state of each discretized cell at each time step during inference. Here we develop Latent Evolution of PDEs (\proj), a simple, fast and scalable method to accelerate the simulation and inverse optimization of PDEs. \proj learns a compact, global representation of the system and efficiently evolves it fully in the latent space with learned evolution models. \proj achieves speed-up by having a much smaller latent dimension to update during long rollout as compared to updating in the input space. We introduce new learning objectives to effectively learn such latent dynamics to ensure long-term stability. We further introduce techniques for speeding up inverse optimization of boundary conditions for PDEs via backpropagation through time in latent space, and an annealing technique to address the non-differentiability and sparse interaction of boundary conditions. We test our method in a 1D benchmark of nonlinear PDEs, 2D  Navier-Stokes flows into turbulent phase and an inverse optimization of boundary conditions in 2D Navier-Stokes flow. Compared to %state-of-the-art deep learning-based surrogate models and 
other strong baselines, we demonstrate up to 128$\times$ reduction in the dimensions to update, and up to 15$\times$ improvement in speed, while achieving competitive accuracy.
\footnote{Project website and code can be found at \url{http://snap.stanford.edu/le_pde/}.}.

\end{abstract}

\section{Introduction}
\label{sec:introduction}

Many problems across science and engineering are described by Partial Differential Equations (PDEs). Among them, temporal PDEs are of huge importance. They describe how the state of a (complex) system evolves with time, and numerically evolving such equations are used for forward prediction and inverse optimization across many disciplines. Example application includes weather forecasting \cite{lynch2008origins}, jet engine design \cite{athanasopoulos2009parametric}, nuclear fusion \cite{carpanese2021development}, laser-plasma interaction \cite{sircombe2006kinetic}, astronomical simulation \cite{courant1967partial}, and molecular modeling \cite{lelievre2016partial}. 

To numerically evolve such PDEs, decades of works have yielded (classical) PDE solvers that are tailored to each specific problem domain \cite{brandstetter2022message}.  Albeit principled and accurate, classical PDE solvers are typically slow due to the small time steps or implicit method required for numerical stability, and their time complexity typically scales linearly or super-linearly with the number of cells the domain is discretized into \cite{keyes2006implicit}. For practical problems in science and engineering, the number of cells at each time step can easily go up to millions or billions and may even require massively parallel supercomputing resources \cite{dubois2008cosmological,chatelain2008billion}. Besides forward modeling, inverse problems, such as inverse optimization of system parameters and inverse parameter inference, also share similar scaling challenge \cite{biegler2003large}.  How to effectively speed up the simulation while maintaining reasonable accuracy remains an important open problem.

Recently, deep learning-based surrogate models have emerged as attractive alternative to complement \cite{um2020solver} or replace classical solvers \cite{sanchez2020learning,li2021fourier}. They directly learn the dynamics from data and alleviate much engineering effort. They typically offer speed-up due to explicit forward mapping \cite{tang2020deep,wu2022learning}, larger time intervals \cite{li2021fourier}, or modeling on a coarser grid \cite{um2020solver,kochkov2021machine}. However, their evolution scales with the discretization, since they typically need to update the state of each discretized cell at each time step, due to the local nature of PDEs \cite{sanchez2020learning2}. For example, if a problem is discretized into 1 million cells, deep learning-based surrogate models (\textit{e.g.}, CNNs, Graph Networks, Neural Operators) will need to evolve these 1 million cells per time step. How to go beyond updating each individual cells and further speed up such models remains a challenge.

\begin{figure}[t]
\begin{center}
\includegraphics[width=1\columnwidth]{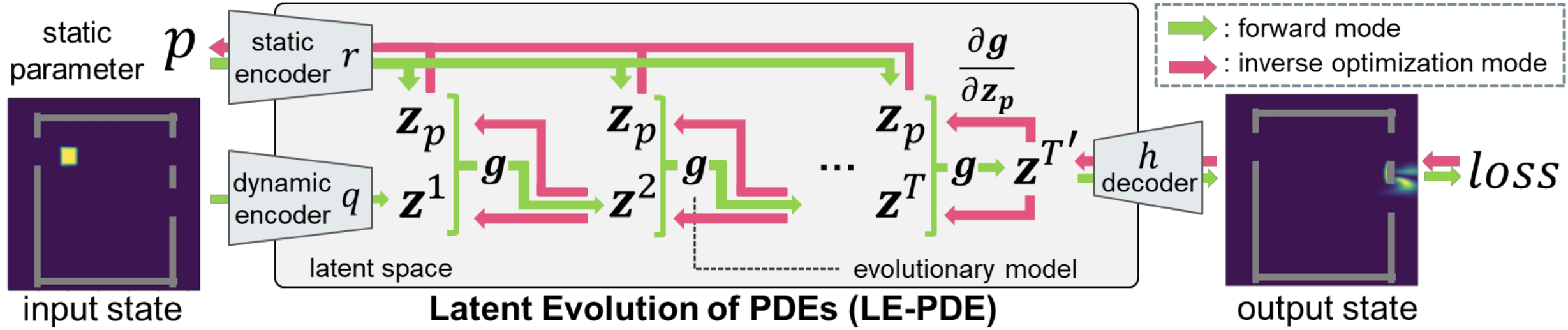}
\end{center}
\caption{\proj schematic. In forward mode (green), \proj evolves the dynamics in a global latent space. In inverse optimization mode (red), it optimizes parameter $p$ (\textit{e.g.} boundary) through latent unrolling. The compressed latent vector and dynamics can significantly speed up both modes.
}
\label{fig:schematic}
\vskip -0.248in
\end{figure}

Here we present Latent Evolution of PDEs (\proj) (Fig. \ref{fig:schematic}), a simple, fast and scalable method to accelerate the simulation and inverse optimization of PDEs. Our key insight is that a common feature of the dynamics of many systems of interest is the presence of dominant, low-dimensional coherent structures, suggesting the possibility of efficiently evolving the system in a low-dimensional global latent space. Based on this observation, we develop \proj, which learns the evolution of dynamics in a global latent space. Here by ``global'' we mean that the dimension of the latent state is fixed, instead of scaling linearly with the number of cells as in local models. \proj consists of a dynamic encoder that compresses the input state into a dynamic latent vector, a static encoder that encodes boundary conditions and equation parameters into a static latent vector, and a latent evolution model that evolves the dynamic latent vector fully in the latent space, and decode via a decoder only as needed. Although the idea of latent evolution has appeared in other domains, such as in computer vision \cite{watters2017visual,van2018relational,udrescu2021symbolic} and robotics \cite{gelada2019deepmdp,hafner2019learning,julian2020scaling,lee2020stochastic}, these domains typically have clear object structure in visual inputs allowing compact representation. PDEs, on the other hand, model dynamics of continuum (\textit{e.g.}, fluids, materials) with infinite dimensions, without a clear object structure, and sometimes with chaotic turbulent dynamics, and it is pivotal to model their long-term evolution accurately. Thus, learning the latent dynamics of PDEs presents unique challenges.  

We introduce a multi-step latent consistency objective, to encourage learning more stable long-term evolution in latent space. Together with the multi-step loss in the input space, they encourage more accurate long-term prediction. To accelerate inverse optimization of PDEs which is pivotal in engineering (\textit{e.g.} optimize the boundary condition so that the evolution of the dynamics optimizes certain predefined objective), we show that \proj can allow faster optimization, via backpropagation through time in latent space instead of in input space. 
To address the challenge that the boundary condition may be non-differentiable or too sparse to receive any gradient, we design an annealing technique for the boundary mask during inverse optimization.

We demonstrate our \proj in standard PDE-learning benchmarks of a 1D family of nonlinear PDEs and a 2D Navier-Stokes flow into turbulent phase, and design an inverse optimization problem in 2D Navier-Stokes flow to probe its capability.
Compared with state-of-the-art deep learning-based surrogate models and other strong baselines, we show up to 128$\times$ reduction in the dimensions to update and up to 15$\times$ speed-up compared to modeling in input space, and competitive accuracy.

\section{Problem Setting and Related Work}
\label{sec:problem_setting}
We consider temporal Partial Differential Equations (PDEs) w.r.t. time $t\in[0,T]$ and multiple spatial dimensions $\x=[x_1,x_2,...x_D]\in\X\subseteq \R^D$. We follow similar notation as in \cite{brandstetter2022message}. Specifically,

\vskip -0.2in
\begin{align}
&\partial_t \u=F(\x,\a,\u,\partial_\x \u, \partial_{\x\x}\u,...), \ \ \ \ \ \ \ \ \ \ \ \ \  \ \ \ \ (t,\x)\in [0,T]\times \mathbb{X}\label{eq:pde},\\
&\u(0,x)=\u^0(\x),\  B[\u](t,\x)=0, \ \ \ \  \ \ \ \  \ \ \ \ \ \ \  \x\in\mathbb{X}, (t,\x)\in [0,T]\times \partial\mathbb{X}.\label{eq:condition}
\end{align}
Here $\u: [0,T]\times \X\to \R^n$ is the solution, which is an infinite-dimensional function. $\a$ is time-independent static parameters of the system, which can be defined on each location $\x$, \textit{e.g.} diffusion coefficient that varies in space but static in time, or a global parameter. $F$ is a linear or nonlinear function on the arguments of $(\x,\a,\u,\partial_\x \u, \partial_{\x\x}\u,...)$. Note that in this work we consider time-independent PDEs where $F$ does not explicitly depend on $t$. $\u^0(\x)$ is the initial condition, and $B[\u](t,\x)=0$ is the boundary condition when $\x$ is on the boundary of the domain $\partial \X$ across all time  $t\in[0,T]$. Here $\partial_\x\u=\frac{\partial\u}{\partial\x}$, $\partial_{\x\x}\u=\frac{\partial^2\u}{\partial\x^2}$ are first- and second-order partial derivatives, which are a matrix and a 3-order tensor, respectively (since $\x$ is a vector). \emph{Solving} such temporal PDEs means computing the state $\u(t,\x)$ for any time $t\in[0,T]$ and location $\x\in\X$ given the above initial and boundary conditions.

\xhdr{Classical solvers for solving PDEs} To numerically solve the above PDEs, classical numerical solvers typically discretize the domain $\X$ into a finite grid or mesh $X=\{c_i\}_{i=1}^N$ with $N$ non-overlapping cells. Then the infinite-dimensional solution function of $\u(t,\x)$ is discretized into $U^k=\{\u_i^k\}_{i=1}^N\in \U$ for each cell $i$ and time $t=t_k,k=1,2,...K$. $\a$ is similarly discretized into $\{a_i\}_{i=1}^N$ with values in each cell. Mainstream numerical methods, including Finite Difference Method (FDM) and Finite Volume Method (FVM),  proceed to evolve such temporal PDEs by solving the equation at state $\{\u_i^{k+1}\}$ at time $t=t_{k+1}$ from state $\{\u_i^{k}\}$ at time $t_k$. These solvers are typically slow due to small time/space intervals required for numerical stability, and needing to update each cell at each time steps. For more detailed information on classical solvers, see Appendix \ref{app:pde_solvers}

\xhdr{Deep learning-based surrogate modeling}
There are two main approaches in deep learning-based surrogate modeling. The first class of method is autoregressive methods, which learns the mapping $f_\theta$ with parameter $\theta$ of the discretized states $U^k$ between consecutive time steps $t_k$ and $t_{k+1}$: $\hat{U}^{k+1}=f_\theta(\hat{U}^k, p), k=0,1,2,...$. Here $\hat{U^k}=\{\hat{\u}_i^k\}_{i=1}^N$ is the model $f_\theta$'s predicted state for $U^k=\{\u_i^k\}_{i=1}^N$ at time $t_k$, with $\hat{U^0}:=U^0$. $p=(\partial \X, \{a_i\}_{i=1}^N)$ is the system parameter which includes the boundary domain $\partial \X$ and discretized static parameters $\{a_i\}_{i=1}^N$. Repetitively apply $f_\theta$ at inference time results in autoregressive rollout

\vskip -0.2in
\begin{equation}
\label{eq:auto_ori}
\hat{U}^{k+m} = \left(f_\theta(\cdot,p)\right)^{(m)}(\hat{U}^k), m\ge 1.
\end{equation}
\vskip -0.1in

Here $f_\theta(\cdot,p):\U\to \U$ is a partial function whose second argument is fulfilled by the static system parameter $p$. Typically $f_\theta$ is modeled using CNNs (if the domain $\X$ is discretized into a grid), Graph Neural Networks (GNNs, if the domain $\X$ is discretized into a mesh). These methods all involve local computation, where the value $u_i^{k+1}$ at cell $i$  at time $t_{k+1}$ depend on its neighbors $\{u_j^k\}_{j\in\mathcal{N}(i)}$ at time $t_k$, where $\mathcal{N}(i)$ is the set of neighbors up to certain hops. Such formulation includes CNN-based models \cite{wang2020towards}, GNN-based models \cite{brandstetter2022message,pfaff2021learning,li2022graph} and their hierarchical counterparts  \cite{sanchez2020learning2,cciccek20163d}. The surrogate modeling with local dynamics makes sense, since the underlying PDE is essentially a local equation that stipulates how the solution function $\u$'s value at location $\x$ depends on the values at its infinitesimal neighborhood. The second class of method is Neural Operators \cite{li2021fourier,raissi2018deep,zhu2018bayesian,bhatnagar2019prediction,khoo2021solving,li2020neural,li2020multipole,lulearning}, which learns a neural network (NN) that approximates a mapping between infinite-dimensional functions. Although having the advantage that the learned mapping is discretization invariant, given a specific discretization, Neural Operators still needs to update the state at each cell based on neighboring cells (and potentially cells far away), which is still inefficient at inference time, especially dealing with larger-scale problems. 
In contrast to the above classes of deep learning-based approaches that both requires local evolution at inference time, our \proj method focuses on improving efficiency. Using a learned global latent space, \proj removes the need for local evolution and can directly evolve the system dynamics via a global latent vectors $\z^k\in \R^{d_z}$ for time $t_k$. This offers great potential for speed-up due to the significant reduction in representation.

\xhdr{Inverse optimization}
Inverse optimization is the problem of  optimizing the parameters $p$ of the PDE, including boundary $\partial\X$ or static parameter $\a$ of the equation, so that a predefined objective  $L_d[\a,\partial\X]=\int_{t=t_s}^{t_e}\ell_d[\u(t,\x)]dt$ is minimized.
Here the state $\u(t,\x)$ implicitly depends on $\a,\partial\X$ through the PDE (Eq. \ref{eq:pde}) and the boundary condition (Eq. \ref{eq:condition}). Such problems have huge importance in engineering, \textit{e.g.} in designing jet engines \cite{athanasopoulos2009parametric} and materials \cite{butler2016computational} where the objective can be minimizing drag or maximizing durability, and inverse parameter inference (\textit{i.e.} history matching) \cite{vernon2014galaxy,williamson2013history,oliver2011recent} where the objective can be maximum a posteriori estimation. To solve such problem, classical methods include adjoint method \cite{talagrand1987variational,tromp2005seismic}, shooting method \cite{keller1976numerical}, collocation method \cite{betts1998survey}, etc. One recent work \cite{allen2022physical} explores optimization via backpropagation through differential physics in the input space, demonstrating speed-up and improved accuracy compared to classical CEM method \cite{rubinstein2004cross}. However, for long rollout and large input size, the computation becomes intensive to the point of needing to save gradients in files. In comparison, \proj allows backpropagation in latent space, and due to the much smaller latent dimension and evolution model, it can significantly reduce the time complexity in inverse optimization.

\xhdr{Reduced-order modeling} A related class of work is reduced-order modeling. Past efforts typically use linear projection into certain basis functions \cite{treuille2006model,berkooz1993proper,gupta2007legendre,wicke2009modular,long2009real,de2012fluid,kim2013subspace,liu2015model} which may not have enough representation power. A few recent works explore NN-based encoding \cite{radford2015unsupervised,wiewel2019latent,kim2019deep,lee2020model,wiewel2020latent,vlachas2022multiscale} for fluid modeling. Compared to the above works, we focus on speeding up simulation and inverse optimization of more general PDEs using expressive NNs, with novel objectives, and demonstrate competitive performance compared to state-of-the-art deep learning-based models for PDEs.

\section{Our approach \proj}
\label{sec:method}
In this section, we detail our Latent Evolution of Partial Differential Equations (\proj) method. We first introduce the model architecture (Sec. \ref{sec:model_architecture}, and then we introduce learning objective to effectively learn faithfully long-term evolution (Sec. \ref{sec:objective}). 
In Sec. \ref{sec:inverse_optimization}, we introduce efficient inverse optimization in latent space endowed by our method.

\subsection{Model architecture}
\label{sec:model_architecture}

The model architecture of \proj consists of four components: (1) a dynamic encoder $q:\U\to \R^{d_z}$ that maps the input state $U^k=\{\u_i^k\}_{i=1}^N\in \U$ to a latent vector $\z^k\in \R^{d_z}$; (2) an (optional) static encoder $r:\mathbb{P}\to \R^{d_{zp}}$ that maps the (optional) system parameter $p\in\mathbb{P}$ to a static latent embedding $\z_p$; (3) a decoder $h:\R^{d_z}\to \U$ that maps the latent vector $\z^k\in \R^{d_z}$ back to the input state $U^k$; (4) a latent evolution model $g:\R^{d_z}\times \R^{d_{zp}}\to \R^{d_z}$ that maps $\z^k\in \R^{d_z}$ at time $t_k$ and static latent embedding $\z_p\in \R^{d_{zp}}$ to $\z^{k+1}\in \R^{d_z}$ at time $t_{k+1}$. We employ the temporal bundling trick \cite{brandstetter2022message} where each input state $U^k$ can include states over a fixed length $S$ of consecutive time steps, in which case each latent vector $\z_k$ will encode such bundle of states, and each latent evolution will predict the latent vector for the next bundle of $S$ steps. $S$ is a hyperparameter and may be chosen depending on the problem, and $S=1$ reduces to no bundling. A schematic of the model architecture and its inference is illustrated in Fig. \ref{fig:schematic}. Importantly, we require that for the dynamic encoder $q$, it needs to have a flatten operation and MultiLayer Perception (MLP) head that maps the feature map into a single fixed-length vector $\z\in R^{d_z}$. In this way, the dimension of the latent space does not scale linearly with the dimension of the input, which has the potential to significantly compress the input, and can make the long-term prediction much more efficient. At inference time, \proj performs autoregressive rollout in latent space $\R^{d_z}$:

\vskip -0.2in
\begin{equation}
\label{eq:latent_evolve}
\hat{U}^{k+m}=h\circ g\left(\cdot,r(p)\right)^{(m)} \circ q (\hat{U}^k)\equiv h\bigg(\underbrace{g(\cdot,r(p))\circ...\circ g(\cdot,r(p))}_{\text{composing}\ m\ \text{times}}\left(q(\hat{U}^k)\right)\bigg).
\end{equation}
\vskip -0.1in

Compared to autoregressive rollout in input space (Eq. \ref{eq:auto_ori}), \proj can significantly improve efficiency with a much smaller dimension of $\z^k\in \R^{d_z}$ compared to $U^k\in \U$. Here we do not limit the architecture for encoder, decoder and latent evolution models. Depending on the input $U^k$, the encoder $q$ and decoder $h$ can be a CNN or GNN with a (required) MLP head. In this work, we focus on input that is discretized as grid, so the encoder and decoder are both CNN+MLP, and leave other architecture (\textit{e.g.} GNN+MLP) for future work. For static encoder $r$, it can be a simple MLP if the system parameter $p$ is a vector (\textit{e.g.} equation parameters) or CNN+MLP if $p$ is a 2D or 3D tensor (\textit{e.g.} boundary mask, spatially varying diffusion coefficient). We model the latent evolution model $g$ as an MLP with residual connection from input to output. The architectures used in our experiments, are detailed in Appendix \ref{app:model_architecture}, together with discussion of its current limitations.

\subsection{Learning objective}
\label{sec:objective}
Learning surrogate models that can faithfully  roll out long-term is an important challenge. Given discretized inputs $\{U^k\},k=1,...K+M$, we introduce the following objective to address it:
\vskip -0.2in
\begin{equation}
\label{eq:objective}
L=\frac{1}{K}\sum_{k=1}^K (L^k_{\text{multi-step}} +  L^k_{\text{recons}} +  L^k_{\text{consistency}}).\\
\end{equation}
\begin{equation}
\label{eq:obj}
\text{where}\,\left\{
\begin{array}{ll}
L_{\text{multi-step}}^k=\sum_{m=1}^M \alpha_m \ell(\hat{U}^{k+m},U^{k+m}),\\
L_{\text{recons}}^k=\ell(h( q(U^k)),U^k)\\
L_{\text{consistency}}^k=\sum_{m=1}^M\frac{||g(\cdot,r(p))^{(m)}\circ q(U^k)-q(U^{k+m})||_2^2}{|| q(U^{k+m})||_2^2}
\end{array}
\right.
\end{equation}
\vskip -0.1in

Here $\ell$ is the loss function for individual predictions, which can typically be MSE or L2 loss. $\hat{U}^{k+m}$ is given in Eq. (\ref{eq:latent_evolve}).  $L_{\text{recons}}^k$ aims to reduce reconstruction loss. $L_{\text{multi-step}}^k$ performs latent multi-step evolution given in Eq. (\ref{eq:latent_evolve}) and compare with the target $U^{k+m}$ in \emph{input} space, up to time horizon $M$. $\alpha_m$ are weights for each time step, which we find that $(\alpha_1,\alpha_2, ...\alpha_M)=(1,0.1,0.1,...0.1)$ works well. Besides encouraging better prediction in input space via $L_\text{multi-step}^k$, we also want a stable long-term rollout in latent space. This is because in inference time, we want to mainly perform autoregressive rollout in latent space, and decode to input space only when needed. Thus, we introduce a novel latent consistency loss $L_{\text{consistency}}^k$, which compares the $m$-step latent rollout $g\left(\cdot,r(p)\right)^{(m)}\circ q(U^k)$ with the latent target $q(U^{k+m})$ in \emph{latent} space. The denominator $|| q(U^{k+m})||_2^2$ serves as normalization to prevent the trivial solution that the latent space collapses to a single point. Taken together, the three terms encourage a more accurate and consistent long-term evolution both in latent and input space. In Sec. \ref{sec:ablation} we will investigate the influence of $L_\text{consistency}^k$ and $L_\text{multi-step}^k$.

\subsection{Accelerating inverse optimization}
\label{sec:inverse_optimization}

In addition to improved efficiency for forward simulation, \proj also allows more efficient inverse optimization, via backpropagation through time (BPTT) in latent space. Given a specified objective  $L_d[p]=\sum_{k=k_s}^{k_e} \ell(U^k)$ which is a discretized version of $L_d[\a,\partial\X]$ in Sec. \ref{sec:problem_setting}, we define the objective: 
\vskip -0.2in
\begin{equation}
\label{eq:inverse_dis}
L_d[p]=\sum_{m=k_s}^{k_e} \ell_d(\hat{U}^m(p))
\end{equation}
\vskip -0.1in

where $\hat{U}^m=\hat{U}^m(p)$ is given by Eq. (\ref{eq:latent_evolve}) setting $k=0$ using our learned \proj, which starts at initial state of $U^0$, encode it and $p$ into latent space, evolves the dynamics in latent space and decode to $\hat{U}^m$ as needed. The static latent embedding $\z_p=r(p)$ influences the latent evolution at each time step via $g(\cdot,r(p))$.  An approximately optimal parameter $p$ can then be found by computing gradients $\frac{\partial L_d[p]}{\partial p}$, using  optimizers such as Adam \cite{kingma2014adam} (The gradient flow is visualized as the red arrows in Fig. \ref{fig:schematic}). When $p$ is a boundary parameter, \textit{e.g.} location of the boundary segments or obstacles, there is a challenge. Specifically, for CNN encoder $q$, the boundary information is typically provided as a binary mask indicating which cells are outside the simulation domain $\Omega$. The discreteness of the mask prevents the backpropagation of the model. Moreover, the boundary cells may interact sparsely with the bulk, which can lead to vanishing gradient during inverse optimization. To address this, we introduce a function that maps $p$ to a soft boundary mask with temperature, and during inverse optimization, anneal the temperature from high to low. This allows the gradient to pass through mask to $p$, and stronger gradient signal. For more information, see Appendix \ref{app:boundary_interpolation}.

\section{Experiments}
In the experiments, we aim to answer the following questions: (1) Does \proj able to learn accurately the long-term evolution of challenging systems, and compare competitively with state-of-the-art methods? (2) How much can \proj reduce representation dimension and improving speed, especially with larger systems?  (3) Can \proj improve and speed up inverse optimization? For the first and second question, since in general there is a fundamental tradeoff between compression (reduction of dimensions to represent a state) and accuracy \cite{tishby2000information,wu2020phase}, \textit{i.e.} the larger the compression to improve speed, the more lossy the representation is, we will need to sacrifice certain amount of accuracy. Therefore, the goal of \proj is to maintain a reasonable or competitive accuracy (maybe slightly underperform state-of-the-art), while achieving significant compression and speed up. Thus, to answer these two questions, we test \proj in standard benchmarks of a 1D family of nonlinear PDEs  to test its generalization to new system parameters (Sec. \ref{sec:burgurs}), and a 2D Navier-Stokes flow up to turbulent phase (Sec. \ref{sec:2d_nv}). The PDEs in the above scenarios have wide and important application in science and engineering. In each domain, we compare \proj's long-term evolution performance, speed and representation dimension with state-of-the-art deep learning-based surrogate models in the domain. Then we answer question  (3) in Section \ref{sec:inverse_opt_exp}. Finally, in Section \ref{sec:ablation}, we investigate the impact of different components of \proj and important hyperparameters.

\subsection{1D family of nonlinear PDEs}
\label{sec:burgurs}

\xhdr{Data and Experiments} In this section, we test \proj's ability to generalize to unseen equations with different parameters in a given family. We use the 1D benchmark in \cite{brandstetter2022message}, whose PDEs are
\vskip -0.25in
\begin{align}
&\left[\partial_t u + \partial_x(\alpha u^2-\beta \partial_x u+\gamma \partial_{xx}u)\right](t,x)=\delta(t,x)\\
&u(0,x)=\delta(0,x), \ \ \ \ \delta(t,x)=\sum_{j=1}^{J}A_j\text{sin}(\omega_j t+2\pi \ell_j x/L+\phi_j)
\end{align}
\vskip -0.15in

Here the parameter $p=(\alpha,\beta,\gamma)$. The term $\delta$ is a forcing term \cite{bar2019learning} with $J=5, L=16$ and coefficients $A_j$ and $\omega_j$ sampled uniformly from $A_j\sim U[-0.5,0.5]$, $\omega_j\sim U[-0.4,0.4]$, $\ell_j\in\{1,2,3\}$, $\phi_j\sim U[0,2\pi)$. Space is uniformly discretized to $n_x=200$ in $[0,16)$ and time is uniformly discretized to $n_t=250$ points in $[0,4]$. Space and time are further downsampled to resolutions of $(n_t,n_x)\in\{(250,100), (250,50), (250,40)\}$. The $\partial_x(\alpha u^2)$ advection term makes the PDE nonlinear. There are 3 scenarios with increasing difficulty: \textbf{E1:} Burgers' equation without diffusion $p=(1,0,0)$; \textbf{E2}: Burgers' equation with variable diffusion $p=(1,\eta,0)$ where $\eta \in[0,0.2]$; \textbf{E3}: mixed scenario with $p=(\alpha,\beta,\gamma)$ where $\alpha\in[0,3],\beta\in[0,0.4]$ and $\gamma\in[0,1]$. \textbf{E1} tests the model's ability to generalize to new conditions with same equation. \textbf{E2} and \textbf{E3} test the model's ability to generalize to novel parameters of PDE with the same family. We compare \proj with state-of-the-art deep learning-based surrogate models for this dataset, specifically  MP-PDE \cite{brandstetter2022message} (a GNN-based model) and Fourier Neural Operators (FNO) \cite{li2021fourier}. For FNO, we compare with two versions: FNO-RNN is the autoregressive version in Section 5.3 of their paper, trained with autoregressive rollout; FNO-PF is FNO improved with the temporal bundling and push-forward trick as implemented in \cite{brandstetter2022message}. To ensure a fair comparison, our \proj use temporal bundling of $S=25$ time steps as in MP-PDE and FNO-PF. We perform hyperparameter search over latent dimension of $\{64,128\}$ and use the model with best validation performance. In addition, we compare with downsampled ground-truth (WENO5), which uses a classical 5$^\text{th}$-order WENO scheme \cite{shu2003high} and explicit Runge-Kutta 4 solver \cite{runge1895numerische,kutta1901beitrag} to generate  the ground-truth data and downsampled to the specified resolution. For all models, we autoregressively roll out to predict the states starting at step 50 until step 250, and record the accumulated MSE, runtime and representation dimension (the dimension of state to update at each time step). Details of the experiments are given in Appendix \ref{app:1d_exp}.

\begin{table}[t]
\centering
\caption{Performance of models in 1D for scenarios \textbf{E1},\textbf{E2}, \textbf{E3}. Accumulated error $=$ $\frac{1}{n_x}\sum_{t,x}\text{MSE}$. Representation dimension ($=S\times n_x$ here) is the number of dimensions to update at each time step. The bold values represent the best performance for experiments and underline  shows second best.
}
\resizebox{1\textwidth}{!}{%
\begin{tabular}{@{}ll|ccccc|cccc|cc} \toprule
& & \multicolumn{5}{c}{\textbf{Accumulated Error $\downarrow$}} & \multicolumn{4}{c}{\textbf{Runtime [ms]} $\downarrow$} & \multicolumn{2}{c}{\textbf{Representation dim} $\downarrow$}  \\
& $(n_t,n_x)$ &  WENO5  & FNO-RNN & FNO-PF & MP-PDE  & \makecell{\textbf{\proj} \\ \textbf{(ours)}}& \multicolumn{1}{c}{WENO5} & \multicolumn{1}{c}{MP-PDE} & \makecell{\textbf{\proj}  \\ \textbf{full (ours)}} & \makecell{\textbf{\proj} \\ \textbf{evo (ours)}} & \multicolumn{1}{c}{MP-PDE} & \makecell{\textbf{\proj} \\ \textbf{(ours)}}\\  \midrule
\textbf{E1} & $(250,100)$  & 2.02 & 11.93 &  \textbf{0.54} &  1.55 & \underline{1.13}& $1.9\times10^3$& 90 & 20 & \textbf{8} &2500 & \textbf{128}\\
\textbf{E1} & $(250,50)$  & 6.23 &  29.98 & \textbf{0.51} & 1.67 & \underline{1.20} & $1.8\times10^3$ & 80 & 20 &\textbf{8} &1250 & \textbf{128}\\
\textbf{E1} & $(250,40)$  & 9.63 & 10.44 &  \textbf{0.57} & 1.47 & \underline{1.17} & $1.7\times10^3$ & 80 &  20 & \textbf{8}&1000 & \textbf{128}\\ \midrule
\textbf{E2} & $(250,100)$  & \underline{1.19} & 17.09 & 2.53 & 1.58 & \textbf{0.77} & $1.9\times10^3$ & 90 & 20 & \textbf{8}&2500 & \textbf{128}\\
\textbf{E2} & $(250,50)$  & 5.35 & 3.57 & 2.27 & \underline{1.63} & \textbf{1.13} & $1.8\times10^3$ & 90 & 20 &\textbf{8} &1250 & \textbf{128}\\
\textbf{E2} & $(250,40)$  & 8.05 & 3.26 & 2.38 & \underline{1.45} & \textbf{1.03} & $1.7\times10^3$ & 80 &  20 & \textbf{8}& 1000 & \textbf{128}\\ \midrule
\textbf{E3} & $(250,100)$  & 4.71 & 10.16 & 5.69 & \underline{4.26} & \textbf{3.39} & $4.8\times10^3$ & 90 & 19 &\textbf{6} & 2500 & \textbf{64}\\
\textbf{E3} & $(250,50)$  & 11.71 & 14.49 & 5.39 & \textbf{3.74} & \underline{3.82} & $4.5\times10^3$ & 90 & 19 &\textbf{6} &1250 & \textbf{64}\\
\textbf{E3} & $(250,40)$  & 15.94  & 20.90 & 5.98 & \textbf{3.70}  & \underline{3.78} & $4.4\times10^3$ & 90 & 20 & \textbf{8}&1000 & \textbf{128} \\ 
 \bottomrule
\end{tabular}}
\label{tab:1d}
\vskip -0.2in
\end{table}

\begin{figure}[t]
\begin{center}
\includegraphics[width=0.75\columnwidth]{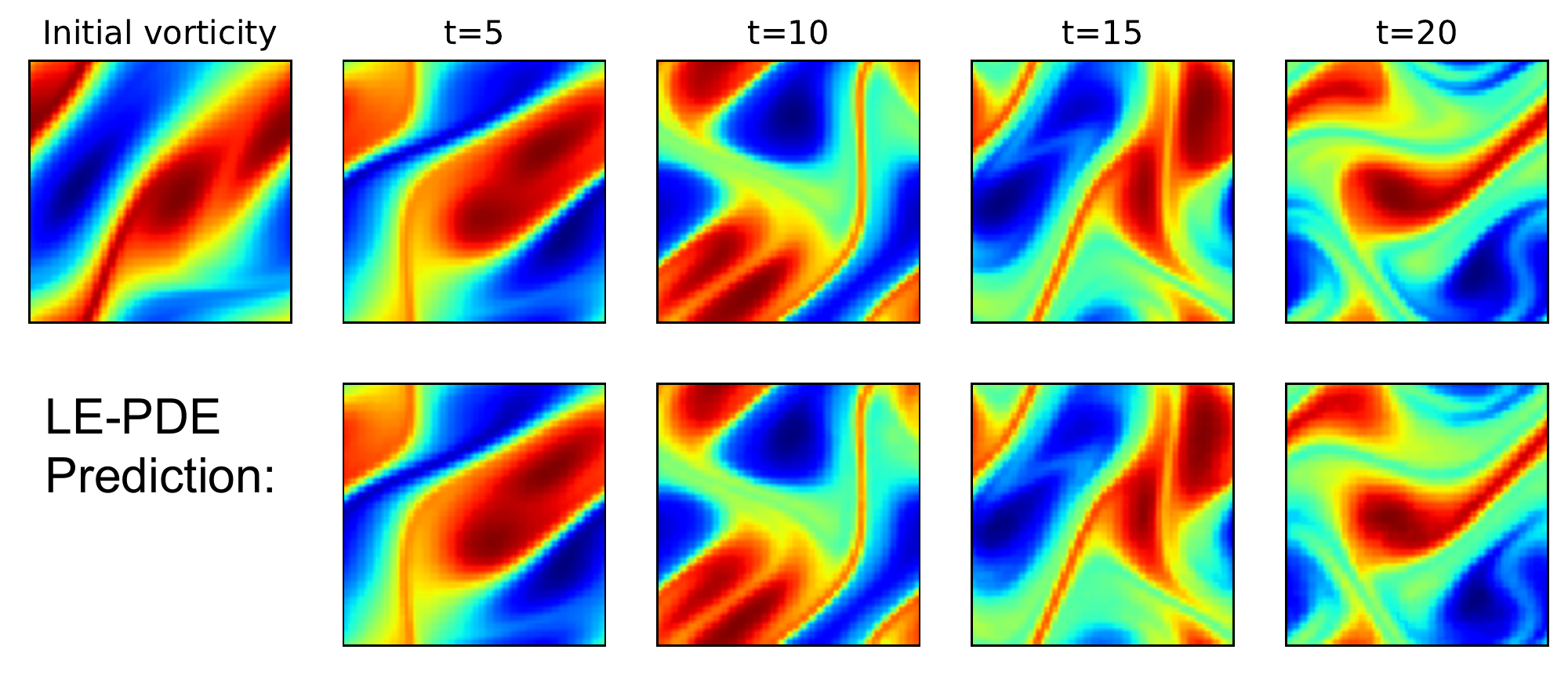}
\vskip -0.2in
\end{center}
\caption{Visualization of rollout for 2D Navier-Stokes PDE ($Re=10^4$), for ground-truth (upper) and \proj (lower, trained with $\nu=10^{-4},N=10^4$). \proj captures detailed dynamics faithfully.}

\label{fig:2d_nv}
\vskip -0.2in
\end{figure}

\xhdr{Results} The result is shown in Table \ref{tab:1d}. We see that since \proj uses 7.8 to 39-fold smaller representation dimension, it achieves significant smaller runtime compared to the MP-PDE model (which is much faster than the classical WENO5 scheme). Here we record the latent evolution time (\proj evo) which is the total time for 200-step latent evolution, and the full time (\proj full), which also includes decoding to the input space at each time step. The time for ``\proj evo'' is relevant when the downstream application is only concerned with state at long-term future (\textit{e.g.} \cite{sircombe2006kinetic}); the time for ``\proj full'' is relevant when every intermediate prediction is also important. \proj achieves up to 15$\times$ speed-up with ``\proj evo'' and 4$\times$ speed-up with ``\proj full''.

With above 7.8$\times$ compression and above 4$\times$ speed-up, \proj still achieves competitive accuracy. For \textbf{E1} scenario, it significantly outperforms both original versions of FNO-RNN and MP-PDE, and only worse than the improved version of FNO-PF. For \textbf{E3}, \proj outperforms both versions of FNO-RNN and FNO-PF, and the performance is on par with  MP-PDE and sometimes better. For \textbf{E2}, \proj outperforms all state-of-the-art models by a large margin. Fig. \ref{fig:1d_rollout} in Appendix \ref{app:1d_exp} shows our model's representative rollout compared to ground-truth. We see that during long-rollout, our model captures the shock formation faithfully. This 1D benchmark shows that  \proj is able to achieve significant speed-up, generalize to novel PDE parameters and achieve competitive long-term rollout.

\subsection{2D Navier-Stokes flow}
\label{sec:2d_nv}

\begin{table}[b!]
\vskip -0.2in
\centering
\caption{Performance of different models in 2D Navier-Stokes flow. 
Runtime is using the $\nu=10^{-3},N=1000$ for predicting 40 steps in the future.
}
\resizebox{1\textwidth}{!}{%
\begin{tabular}{@{}l|ccc|cccc@{}}
\hline
\cline{2-7}
Method       & \makecell{Representation \\ dimensions} & \makecell{Runtime \\ full $[ms]$} & \makecell{Runtime \\ evo $[ms]$} & \makecell{$\nu=10^{-3}$ \\ $T=50$ \\ $N=1000$ } & \makecell{$\nu=10^{-4}$ \\ $T=30$ \\ $N=1000$ } & \makecell{$\nu=10^{-4}$ \\ $T=30$ \\ $N=10000$ }  & \makecell{$\nu=10^{-5}$ \\ $T=20$ \\ $N=1000$ }\\ 
\hline
FNO-3D \cite{li2021fourier}&    4096 & 24 & 24 & 0.0086  & 0.1918  & 0.0820 & 0.1893 \\
FNO-2D \cite{li2021fourier} &     4096 & 140 & 140 & 0.0128  & 0.1559  & 0.0834 & 0.1556 \\
U-Net  \cite{ronneberger2015u}             & 4096   &   813  &   813     & 0.0245  & 0.2051      & 0.1190                & 0.1982 \\
TF-Net \cite{wang2020towards}    & 4096& 428 &  428 & 0.0225    & 0.2253       & 0.1168    & 0.2268 \\
ResNet \cite{he2016deep}    & 4096& 317 & 317 & 0.0701   &  0.2871       &0.2311   & 0.2753 \\
\hline
\textbf{\proj (ours)}   & 256 & 48 & 15 & 0.0146  &     0.1936   & 0.1115  & 0.1862 \\
\Xhline{2\arrayrulewidth}
\end{tabular}}
\label{tab:2d_flow}
\vskip -0.1in
\end{table}

\xhdr{Data and Experiments} We test \proj in a  2D benchmark \cite{li2021fourier} of Navier-Stokes equation. Navier-Stokes equation has wide application science and engineering, including weather forecasting, jet engine design, etc. It becomes more challenging to simulate when entering the turbulent phase, which shows multiscale dynamics and chaotic behavior. Specifically, we test our model in a viscous, incompressible fluid in vorticity form in a unit torus:

\vskip -0.25in
\begin{align}
\partial_t w(t,x)+u(t,x)\cdot \nabla w(t,x)&=\nu \Delta w(t,x)+f(x), \ \ \ \ x\in(0,1)^2,t\in(0,T]\\
\nabla\cdot u(t,x)&=0, \ \ \ \ \ \ \ \ \ \  \ \ \ \ \ \ \ \ \  \ \ \ \ \ \ \ \ \ \ \ \ x\in(0,1)^2,t\in[0,T]\\
w(0,x)&=w_0(x), \ \ \ \ \ \ \ \ \ \ \ \ \ \ \ \ \ \  \ \ \ \  \ x\in(0,1)^2
\end{align}
\vskip -0.1in

Here $w(t,x)=\nabla\times u(t,x)$ is the vorticity, $\nu\in \R_+$ is the viscosity coefficient. The domain is discretized into $64\times 64$ grid. We test with viscosities of $\nu=10^{-3},10^{-4},10^{-5}$. The fluid is turbulent for $\nu=10^{-4},10^{-5}$ ($Re\ge10^{4}$). We compare state-of-the-art learning-based model Fourier Neural Operator (FNO) \cite{li2021fourier} for this problem, and strong baselines of TF-Net \cite{wang2020towards}, U-Net \cite{ronneberger2015u} and ResNet \cite{he2016deep}. For FNO, the FNO-2D performs autoregressive rollout, and FNO-3D directly maps the past 10 steps into all future steps. To ensure a fair comparison, here our \proj uses past 10 steps to predict one future step and temporal bundling $S=1$ (no bundling), the same setting as in FNO-2D. We use relative L2 norm (normalized by ground-truth's L2 norm) as metric, same as in \cite{li2021fourier}.

\xhdr{Results} The results are shown in Table \ref{tab:2d_flow}. Similar to 1D case, \proj is able to compress the representation dimension by 16-fold. Hence, compared with FNO-2D which is also autoregressive, \proj achieves 9.3-fold speed-up with latent evolution and 2.9-fold speed-up with full decoding. Compared with FNO-3D that directly maps all input time steps to all output times steps (which cannot generalize beyond the time range given), \proj's runtime is still  1.6$\times$ faster for latent evolution. For rollout L2 loss, \proj significantly outperforms strong baselines of ResNet and U-Net, and TF-Net which is designed to model turbulent flow. Its performance is on par with FNO-3D with $\nu=10^{-4},N=1000$ and the most difficult $\nu=10^{-5},N=1000$ and slightly underperforms FNO-2D in other scenarios. Fig. \ref{fig:2d_nv} shows the visualization of \proj comparing with ground-truth, under the turbulent $\nu=10^{-4},N=10000$ scenario. We see that \proj captures the detailed dynamics accurately. For more details, see Appendix \ref{app:2d_nv}. To explore how \proj can model and accelerate the simulation of  systems with a larger scale, in Appendix \ref{app:3d_nv} we explore modeling a 3D Navier-Stokes flow with millions of cells per time step, and show more significant speed-up.

\begin{figure}[t]
\centering
\begin{subfigure}{0.49\textwidth}
    \includegraphics[width=\textwidth]{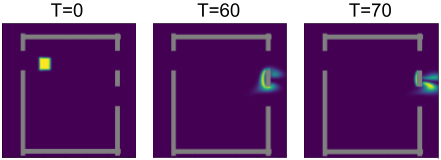}
    \caption{Trajectory generated by ground-truth solver with an initial randomly generated boundary parameter.}
    \label{fig:first}
\end{subfigure}
\hfill\hfill
\begin{subfigure}{0.49\textwidth}
    \includegraphics[width=\textwidth]{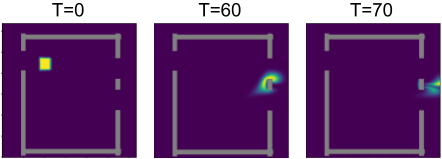}
    \caption{Trajectory generated by ground-truth solver with boundary parameter optimized by \proj.}
    \label{fig:second}
\end{subfigure}
\begin{subfigure}{0.323\textwidth}
    \includegraphics[width=\textwidth]{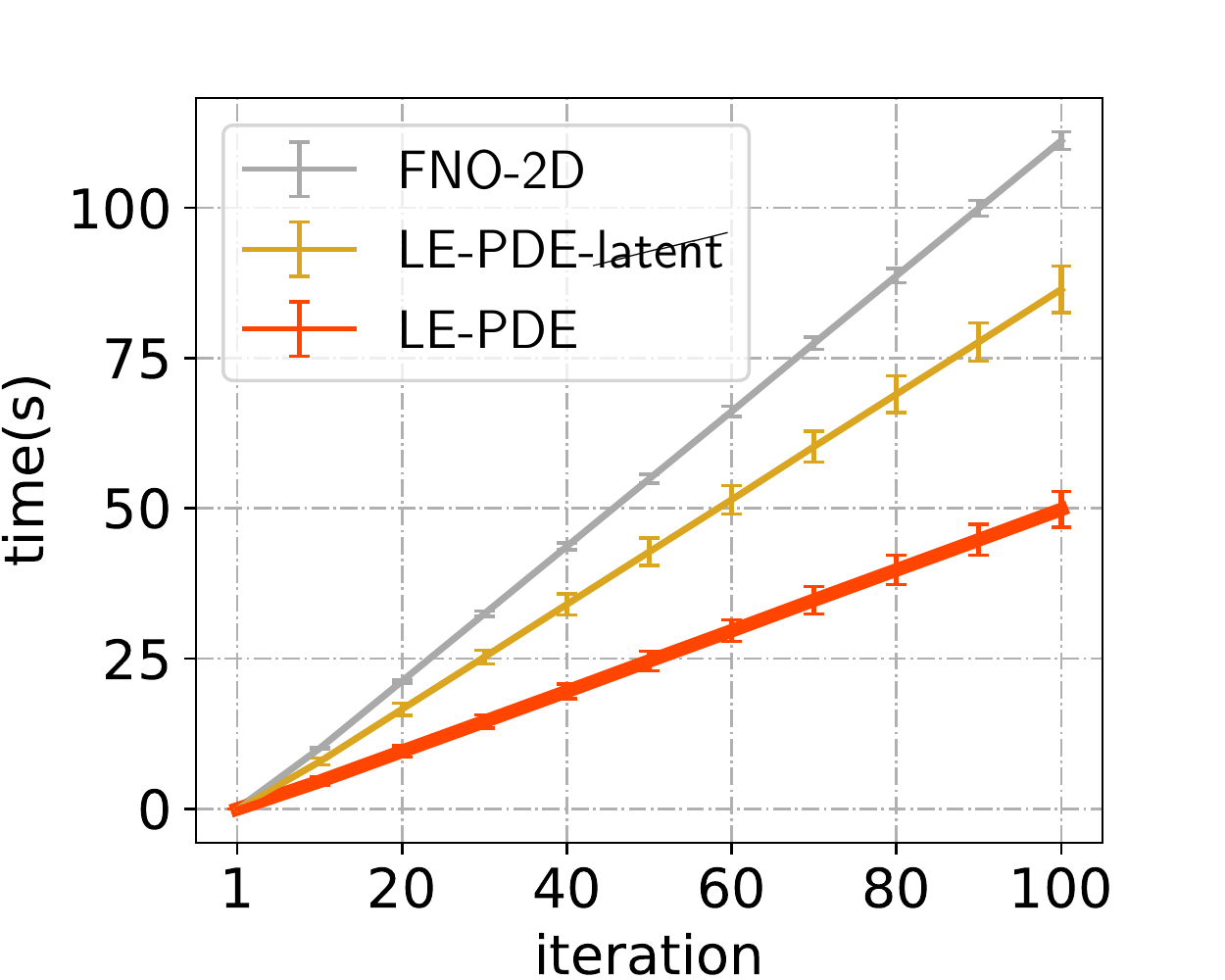}
    \caption{Runtime.}
    \label{fig:third}
\end{subfigure}
\begin{subfigure}{0.323\textwidth}
    \includegraphics[width=\textwidth]{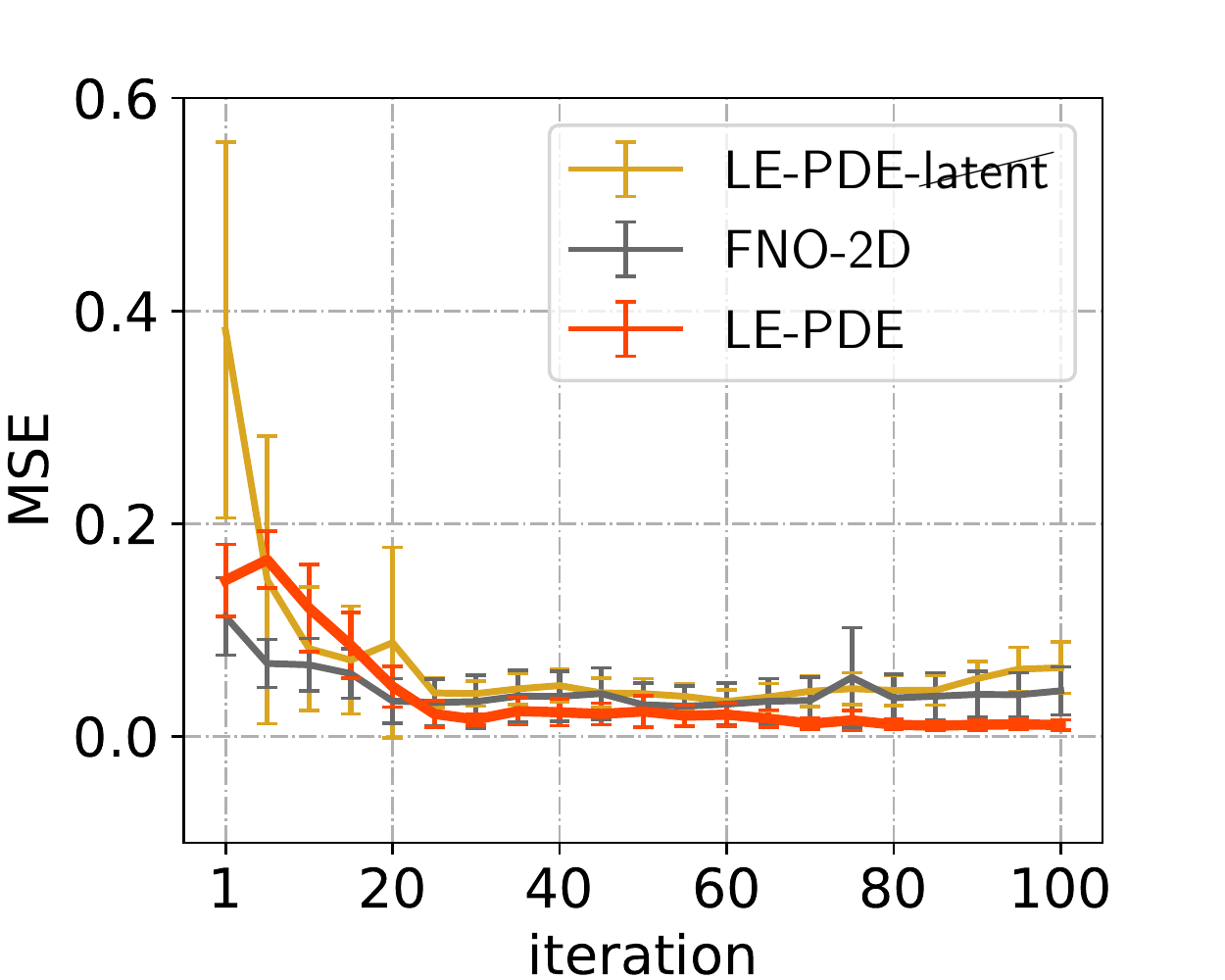}
    \caption{Objective function.}
    \label{fig:fifth}
\end{subfigure}
\begin{subfigure}{0.323\textwidth}
    \includegraphics[width=\textwidth]{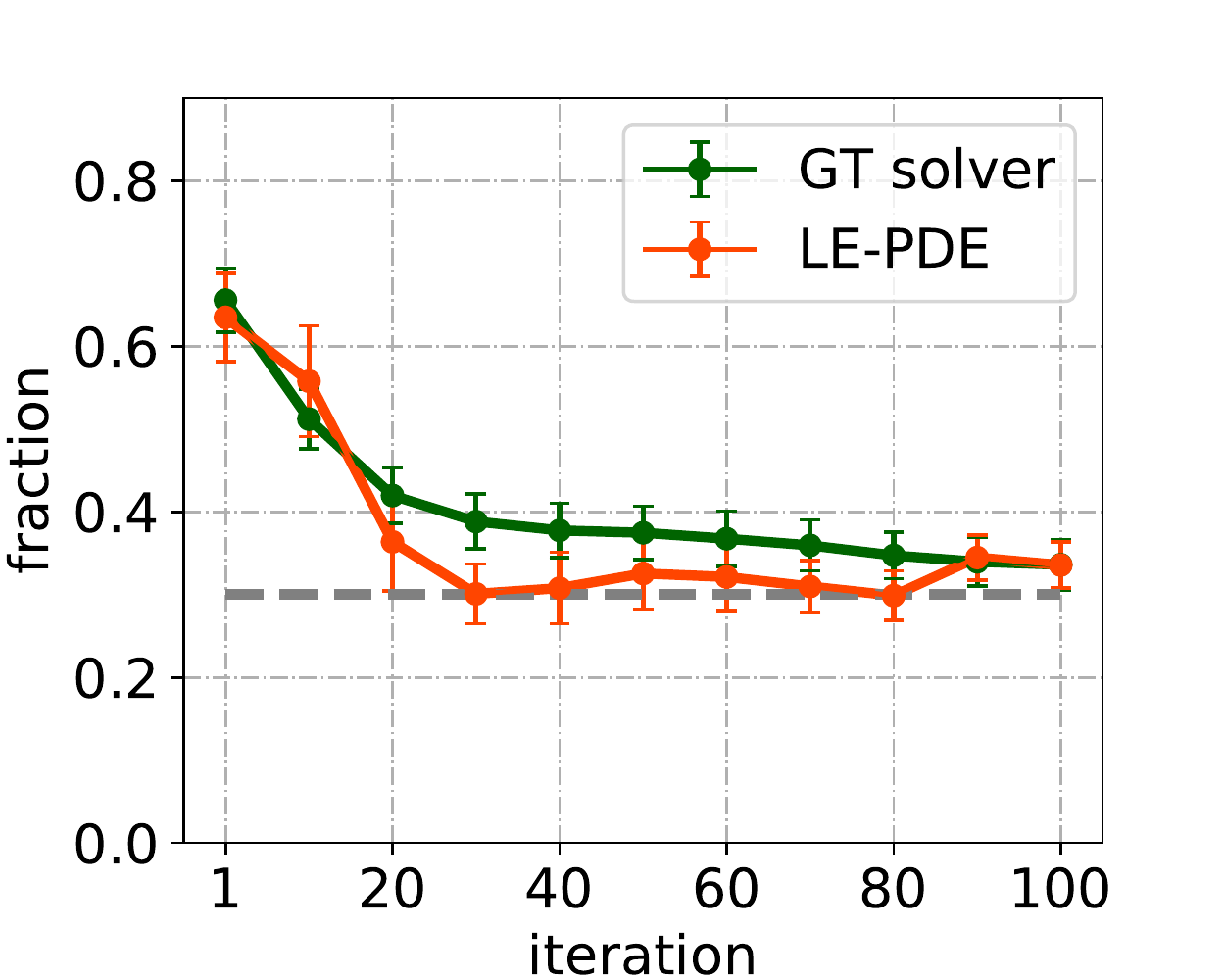}
    \caption{Smoke amount at lower outlet.}
    \label{fig:forth}
\end{subfigure}

\caption{Numerical results associated with inverse optimization of boundary (inlet and outlet designing) in Sec. \ref{sec:inverse_opt_exp}. (a) shows a trajectory generated by ground-truth (GT) solver with an initial randomly generated boundary parameters (y-position of inlet and two outlets), with lower outlet passing 55.18\% of smoke; (b) with optimized boundary parameters, with lower outlet passing 31.79\% of smoke, very near the objective percentage of 30\%. (c) Runtime and (d) learning curve (Eq. \ref{eq:inverse_dis}) for inverse optimization at different iteration steps; (e) For \proj, fraction of smoke passing through the lower outlet computed by GT solver (green) and estimated by \proj (orange).
Error bar denotes 95\% confidence interval over 50 runs with random initial conditions.}
\vskip -0.2in
\label{fig:inv_opt}
\end{figure}

\subsection{Accelerating inverse optimization of boundary conditions}
\label{sec:inverse_opt_exp}

\begin{wraptable}{r}{0.5\textwidth}
\vskip -0.1in
\caption{Comparison of \proj with baselines. %\proj achieves above 1.7$\times$ speed-up and much lower error computed by ground-truth solver.
}
\centering
\resizebox{0.5\textwidth}{!}{
\begin{tabular}{ccc}
\hline
& \makecell{GT-solver Error\\ (Model estimated Error)} & \makecell{Runtime \\ $[s]$}  \\
\hline
\projnolatent & 0.305 (0.123) & 86.42 \\ 
FNO-2D  & 0.124 (0.004) & 111.14 \\
\hline
\textbf{\proj (ours)} & \textbf{0.035} (0.036) & \textbf{49.81}   \\
\hline
\end{tabular}
}
\vskip -0.1in
\label{table:invcomparison}
\end{wraptable}
\textbf{Data and Experiments}. In this subsection, we set out to answer question (3), \textit{i.e.}  Can \proj improve and speed up inverse optimization? We are interested in long time frame scenarios where the pre-defined objective $L_d$ in Eq. (\ref{eq:inverse_dis}) depends on the prediction after long-term rollout. Such problems are challenging and have  implications in engineering, \textit{e.g.} fluid control \cite{mcnamara2004fluid,Holl2020Learning}, laser design for laser-plasma interaction \cite{sircombe2006kinetic} and nuclear fusion \cite{zylstra2022burning}. To evaluate, we build a 2D Navier-Stokes flow in a family of boundary conditions using PhiFlow \cite{phiflow} as our ground-truth solver, shown in Fig. \ref{fig:first}, \ref{fig:second}. Specifically, we create a cubical boundary with one inlet and two outlets on a grid space of size $128^2$. We initialize the velocity and smoke on this domain and advect the dynamics by performing rollout. The objective of the inverse design here is to optimize the boundary parameter $p$, \textit{i.e.} the $y$-locations of the inlet and outlets, so that the amount of smoke passing through the two outlets coincides with pre-specified proportions $0.3$ and $0.7$. This setting is challenging since a slight change in boundary (up to a few cells) can have large influence in long-term rollout and the predefined objective.

As baseline methods, we use our \proj's ablated version without latent evolution (essentially a CNN, which we call \projnolatent) and the FNO-2D \cite{li2021fourier}, both of which update the states in input space, while \proj evolves in a $128$-dimensional latent space (128$\times$ compression). To ensure a fair comparison, all models predict the next step using 1 past step without temporal bundling, and trained with 4-step rollout. We train all models with $400$ generated trajectories of length $100$ and test with $40$ trajectories. After training, we perform inverse optimization w.r.t. the boundary parameter $p$ with the trained models using Eq. \ref{eq:inverse_dis}, starting with 50 initial configurations each with random initial location of smoke and random initial configuration of $p$. For \projnolatent and FNO-2D, they need to backpropagate through 80 steps of rollout in input space as in \cite{allen2022physical,qzhaoInverseProblemGNN}, while \proj backpropagates through 80 steps of latent rollout. Then the optimized boundary parameter is fed to the ground-truth solver for rollout and evaluate. For the optimized parameter, we measure the total amount of smoke simulated by the solver passing through two respective outlets and take their ratio. The evaluation metric is the average ratio across all 50 configurations: see also Appendix \ref{app:inverse}.

\textbf{Results}.\, We observe that \proj improves the overall speed by 73\% compared with \projnolatent and by 123\% compared with FNO-2D (Fig. \ref{fig:third}, Table \ref{table:invcomparison}). The result indicates a corollary of the use of low dimensional representation because Jacobian matrix of evolution operator is reduced to be of smaller size and suppresses the complexity associated with the chain rule to compute gradients of the objective function. While achieving the significant speed-up, the capability of the \proj to design the boundary is also reasonable. Fig. \ref{fig:fifth} shows the loss of the objective function achieved the lowest value while the others are comparably large. The estimated proportion of smoke hit the target fraction $0.3$ at an early stage of design iterations and coincide with the fraction simulated by the ground-truth solver in the end  (Fig. \ref{fig:forth}). %We note that, a
As Table \ref{table:invcomparison} shows, FNO-2D achieves the lowest score in model estimated error from the target fraction 0.3 while its ground-truth solver (GT-solver) error is 30$\times$ larger. This shows ``overfitting'' of the boundary parameter by FNO-2D, \textit{i.e.} the optimized parameter is not sufficiently generalized to work for a ground-truth solver. In this sense, \proj achieved to design the most generalized boundary parameter: the difference between the two errors is the smallest among the others.

\vskip -2.5in
\subsection{Ablation study}
\label{sec:ablation}
\begin{wraptable}{r}{0.4\textwidth}
\vskip -0.15in
\centering
\caption{Error for ablated versions of \proj in 1D  and 2D.}
\resizebox{0.4\textwidth}{!}{
\begin{tabular}{cc|c}
\hline
& \makecell{1D}  & \makecell{2D} \\
\hline
\textbf{\proj (ours)} & \textbf{1.127} &  \textbf{0.1861}     \\
\hline
no $L_\text{multi-step}$ & 3.337 &   0.2156  \\ % update
no $L_\text{consistency}$ & 6.386 &  0.2316  \\ % update
no $L_\text{recons}$ & 1.506 &  0.2025  \\ % update
Time horizon $M=1$ & 5.710 &  0.2860   \\ % update
Time horizon $M=3$ & 1.234 &  0.2010  \\ % update
Time horizon $M=4$ & 1.127 & 0.1861   \\ % update
Time horizon $M=6$ & 1.924 &  0.1923  \\ % update
\hline
\end{tabular}
}
\label{table:ablation}
\vskip -0.05in
\end{wraptable}
In this section, we investigate how each component of our \proj influences the performance. Importantly, we are interested in how each of the three components: multi-step loss $L_\text{multi-step}$, latent consistency loss $L_\text{consistency}$ and reconstruction loss $L_\text{recons}$ contribute to the performance, and how the time horizon $M$ and the latent dimension $d_z$ influence the result. %Furthermore, we compare with a variant (Pretrain with $L_\text{recons}$) where we first pretrain the encoder and decoder with only the reconstruction loss $L_\text{recons}$, then freeze the encoder and decoder and train the latent evolution model $g$ and static encoder $r$ with latent consistency loss $L_\text{consistency}$ only. This variant mimics the techniques in \cite{lee2020model}. 
For dataset, we focus on representative scenarios in 1D  (Sec. \ref{sec:burgurs}) and 2D  (Sec. \ref{sec:2d_nv}), specifically the \textbf{E2} scenario with $(n_t,n_x)=(250,50)$ for 1D, and $(\nu=10^{-5},T=20,N=1000)$ scenario for 2D, which lies at mid- to difficult spectrum of each dataset. We have observed similar trends in other scenarios. From Table \ref{table:ablation}, we see that all three components  $L_\text{multi-step}$, $L_\text{consistency}$ and $L_\text{recons}$ are necessary and pivotal in ensuring a good performance. %The model that pretrain with $L_\text{recons}$ results in a worse performance, because during pretraining the autoencoder, the loss does not know which features to extract to ensure a better long-term rollout. Our \proj, on the other hand, trains all the components simultaneously, so the encoder and decoder is guided by what features to extract for better long-term prediction. 
The time horizon $M$ in the loss is also important. If too short (\textit{e.g.} $M=1$), it does not encourage accurate long-term rollout. Increasing $M$ helps reducing error, but will be countered by less number of examples (since having to leave room for more steps in the future). We find the sweet spot is at $M=4$, which achieves a good tradeoff. In Fig. \ref{fig:ablation} in Appendix \ref{app:ablation}, we show how the error and evolution runtime change with varying size of latent dimension $d_z$. We observe that reduction of runtime with decreasing latent dimension $d_z$, and that the error is lowest at $d_z=64$ for 1D and $d_z=256$ for 2D, suggesting the intrinsic dimension of each problem.

\section{Discussion and Conclusion}
In this work, we have introduced \proj, a simple, fast and scalable method for accelerating simulation and inverse optimization of PDEs, including its simple architecture, objective and inverse optimization techniques. Compared with state-of-the-art deep learning-based surrogate models, we demonstrate that it achieves up to 128 $\times$ reduction in the dimensions to update and up to 15$\times$ improvement in speed, while achieving competitive accuracy. Ablation study shows both multi-step objective and latent-consistency objectives are pivotal in ensuring accurate long-term rollout.
We hope our method will make a useful step in accelerating simulation and inverse optimization of PDEs, pivotal in science and engineering.

\ack{}

We thank Sophia Kivelson, Jacqueline Yau, Rex Ying, Paulo Alves, Frederico Fiuza, Jason Chou, Qingqing Zhao for discussions and for providing feedback on our manuscript.
We also gratefully acknowledge the support of
DARPA under Nos. HR00112190039 (TAMI), N660011924033 (MCS);
ARO under Nos. W911NF-16-1-0342 (MURI), W911NF-16-1-0171 (DURIP);
NSF under Nos. OAC-1835598 (CINES), OAC-1934578 (HDR), CCF-1918940 (Expeditions), 
NIH under No. 3U54HG010426-04S1 (HuBMAP),
Stanford Data Science Initiative, 
Wu Tsai Neurosciences Institute,
Amazon, Docomo, GSK, Hitachi, Intel, JPMorgan Chase, Juniper Networks, KDDI, NEC, and Toshiba.

The content is solely the responsibility of the authors and does not necessarily represent the official views of the funding entities.

\bibliographystyle{IEEEtran}
\bibliography{references}
\section*{Checklist}

\begin{enumerate}

\item For all authors...
\begin{enumerate}
  \item Do the main claims made in the abstract and introduction accurately reflect the paper's contributions and scope?
    \answerYes{}
  \item Did you describe the limitations of your work?
    \answerYes{In Appendix \ref{app:model_architecture}.}
  \item Did you discuss any potential negative societal impacts of your work?
    \answerYes{In Appendix \ref{app:social_impact}}
  \item Have you read the ethics review guidelines and ensured that your paper conforms to them?
    \answerYes{}
\end{enumerate}

\item If you are including theoretical results...
\begin{enumerate}
  \item Did you state the full set of assumptions of all theoretical results?
    \answerNA{}
        \item Did you include complete proofs of all theoretical results?
    \answerNA{}
\end{enumerate}

\item If you ran experiments...
\begin{enumerate}
  \item Did you include the code, data, and instructions needed to reproduce the main experimental results (either in the supplemental material or as a URL)?
    \answerYes{The Appendix includes full details on model architecture, training and evaluation to reproduce the experimental results. Code and data will be released upon publication of the paper.}
  \item Did you specify all the training details (e.g., data splits, hyperparameters, how they were chosen)?
    \answerYes{Important training details are included in main text, and full details to reproduce the experiments are included in the Appendix.}
        \item Did you report error bars (e.g., with respect to the random seed after running experiments multiple times)?
   \answerYes{Yes in Section 4.4. For Sections 4.1 and 4.2, the benchmarks do not include the error bars and we use the same format. }
        \item Did you include the total amount of compute and the type of resources used (e.g., type of GPUs, internal cluster, or cloud provider)?
    \answerYes{In Appendix \ref{app:1d_exp},\ref{app:2d_nv},\ref{app:3d_nv},\ref{app:inverse}.}
\end{enumerate}

\item If you are using existing assets (e.g., code, data, models) or curating/releasing new assets...
\begin{enumerate}
  \item If your work uses existing assets, did you cite the creators?
    \answerYes{In Appendix \ref{app:1d_exp},\ref{app:2d_nv},\ref{app:3d_nv},\ref{app:inverse}.}
  \item Did you mention the license of the assets?
    \answerYes{In Appendix \ref{app:1d_exp},\ref{app:2d_nv},\ref{app:3d_nv},\ref{app:inverse}.}
  \item Did you include any new assets either in the supplemental material or as a URL?
    \answerYes{In Appendix \ref{app:1d_exp},\ref{app:2d_nv},\ref{app:3d_nv},\ref{app:inverse}.}
  \item Did you discuss whether and how consent was obtained from people whose data you're using/curating?
    \answerYes{In Appendix \ref{app:1d_exp},\ref{app:2d_nv},\ref{app:3d_nv},\ref{app:inverse}.}
  \item Did you discuss whether the data you are using/curating contains personally identifiable information or offensive content?
    \answerYes{In Appendix \ref{app:1d_exp},\ref{app:2d_nv},\ref{app:3d_nv},\ref{app:inverse}.}
\end{enumerate}

\item If you used crowdsourcing or conducted research with human subjects...
\begin{enumerate}
  \item Did you include the full text of instructions given to participants and screenshots, if applicable?
    \answerNA{}
  \item Did you describe any potential participant risks, with links to Institutional Review Board (IRB) approvals, if applicable?
    \answerNA{}
  \item Did you include the estimated hourly wage paid to participants and the total amount spent on participant compensation?
    \answerNA{}
\end{enumerate}

\end{enumerate}

\newpage
\appendix
\begin{center}
\begin{huge}
\textbf{Appendix}
\end{huge}
\end{center}

\makeatletter 
\renewcommand{\thefigure}{S\arabic{figure}}
\makeatother
\setcounter{figure}{0}
\begin{figure}[h]
\centering
\begin{subfigure}{0.47\textwidth}
    \includegraphics[width=\textwidth]{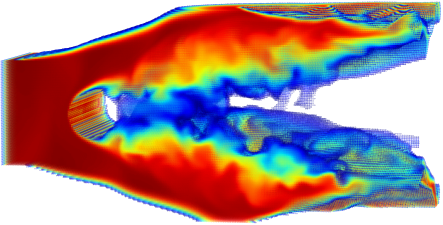}
    \caption{Initial condition at $t=0$}
    \label{fig:3d_y_t=0}
\end{subfigure}

\begin{subfigure}{0.47\textwidth}
    \includegraphics[width=\textwidth]{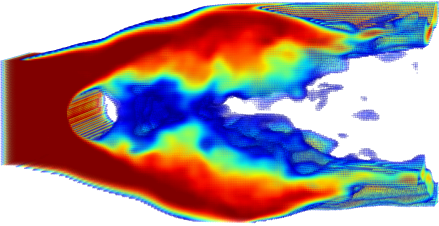}
    \caption{\proj  prediction at $t=20$}
    \label{fig:3d_pred_t=10}
\end{subfigure}
\hfill
\begin{subfigure}{0.47\textwidth}
    \includegraphics[width=\textwidth]{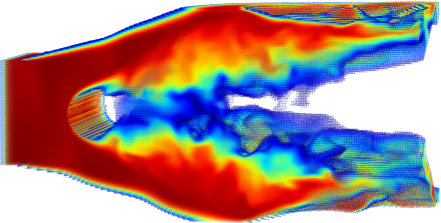}
    \caption{Ground-truth at $t=20$}
    \label{fig:3d_y_t=10}
\end{subfigure}
\begin{subfigure}{0.47\textwidth}
    \includegraphics[width=\textwidth]{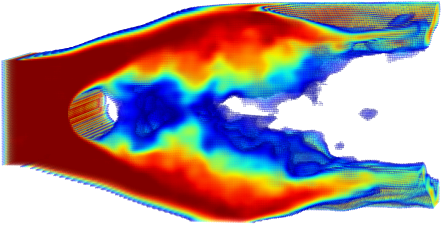}
    \caption{\proj  prediction at $t=40$}
    \label{fig:3d_pred_t=20}
\end{subfigure}
\hfill
\begin{subfigure}{0.47\textwidth}
    \includegraphics[width=\textwidth]{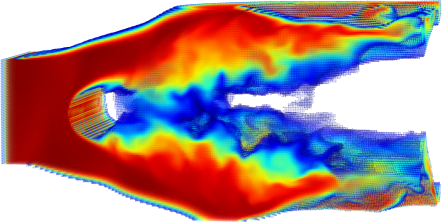}
    \caption{Ground-truth at $t=40$}
    \label{fig:3d_y_t=20}
\end{subfigure}
\caption{Visualization of \proj testing on predicting the dynamics of turbulent 3D Navier-Stokes flow through the cylinder with a novel Reynolds number (detail in Appendix \ref{app:3d_nv}). The input domain of size $2\times1\times1$ is discretized into a 3D grid of $256\times128\times128$, resulting in 4.19 million cells per time step. \textbf{Compression:} \proj learns latent dynamics with latent dimension of $d_z=128$, achieving a \textbf{130,000}$\times$ reduction in representation dimension, compared with 4.19 million cells times 4 features per cell ($\rho,v_x,v_y,v_z$) in input space. \textbf{Prediction quality:} The visualization is shown at a cross-section of $x=50/128\times1$ along the direction of the cylinder. We see that compared with ground-truth (c)(e), \proj (b)(d) captures the turbulent dynamics reasonably well, predicting both high-level and low-level dynamics in a qualitatively faithful way. This shows the scalability of \proj to large-scale simulations of PDE. \textbf{Speed-up:} To predict the state at $t=40$, on an Nvidia Quadro RTX 8000 48GB GPU, the ground-truth solver PhiFlow \cite{phiflow} uses 70.80s, an ablation our \projnolatent without latent evolution (essentially a CNN) takes 1.03s, while our \proj takes only 0.084s. \proj achieves an \textbf{840}$\times$ speed-up compared to the ground-truth solver, and \textbf{12.3$\times$} speed-up compared to the ablation model without latent evolution.}
\label{fig:3d_nv}
\end{figure}
\vskip 0.1in

In the Appendix, we provide details that complement the main text. In Appendix \ref{app:pde_solvers}, we give a brief introduction to classical solvers. In Appendix \ref{app:boundary_interpolation}, we explain details about boundary interpolation and annealing technique used in Section \ref{sec:inverse_opt_exp}. In Appendix \ref{app:model_architecture}, we give full explanation on the architecture of \proj used throughout the experiments. The following three sections explain details on parameter settings of experiments: 1D family of nonlinear PDEs (Appendix \ref{app:1d_exp}), 2D Navier-Stokes flow (Appendix \ref{app:2d_nv}) and 3D Navier-Stokes flow (Appendix \ref{app:3d_nv}). In appendix \ref{app:inverse}, we give details of boundary inverse optimization conducted in Section \ref{sec:inverse_opt_exp}. In appendix \ref{app:ablation}, we show ablation study for \proj's important parameters. In Appendix \ref{app:social_impact}, we discuss the broader social impact of our work. In appendix \ref{app:pareto}, we give comparison of trade-off between some metrics for \proj and some strong baselines. In addition, in Appendix \ref{app:lfm}, we compare \proj to another model exploiting latent evolution method from several aspects. We present the influence of varying noise amplitude with some tables in Appendix \ref{sec:abnoise1d}. Finally, in Appendix \ref{app:encoder}, we show the ablation study for various encoders in different scenarios. 

\section{Classical Numerical Solvers for PDEs}
\label{app:pde_solvers}

We refer the readers to  \cite{brandstetter2022message} Section 2.2 and Appendix for a high-level introduction of the classical PDE solvers. One thing that is in common with the Finite Difference Method (FDM), Finite Volume Method (FVM) is that they all need to update the state of each cell at each time step. This stems from that the methods require discretization of the domain $\mathbb{X}$ and solution $\mathbf{u}$ into a grid $X$. For large-systems with millions or billions of cells, it will result in extremely slow simulation, as is also shown in Appendix \ref{app:3d_nv} where a classical solver takes extremely long to evolve a 3D system with millions of cells per time step.

\section{Boundary Interpolation and Annealing Technique}
\label{app:boundary_interpolation}
\xhdr{Boundary interpolation} In order to allow gradients to pass through to the boundary parameter $p$, we introduce a \textit{continuous boundary mask} that continuously interpolates a discrete boundary mask and continuous variables. Here, for the later convenience, we regard a mask as a function from a grid structure $\mathbb{N}_{\leq 128}^{\times 2}$ to $[0, 1]$. Because boundary is composed by $1$-dimensional segments, we use a $1$-dimensional sigmoid function for the interpolation. Specifically, we define a sigmoid-interpolation function on a segment as a map to a real from a natural number $i$ conditioned by a pair of continuous variables $x_{1}, x_{2}$ and positive real $\beta$:
\begin{equation}
\label{eq:sigint}
f(\ i\ |\ x_{1}, x_{2}, \beta) =  \left\{
\begin{array}{ll}
\operatorname{sigmoid}(\frac{i - x_{1}}{\beta}), \ \ \ \ \ \ \ \ \ \ \ \ \ \ \ \ \ \ \ \ \ \ \ \  i \leq x_{1}, \\
\operatorname{sigmoid}(\frac{x_{2} - i}{\beta}), \ \ \ \ \ \ \ \ \ \ \ \ \ \ \ \ \ \ \ \ \ \ \  x_{2} \leq i, \\
\operatorname{sigmoid}(\frac{GM_{-1}(|i - x_{1}|, |i - x_{2}|)}{\beta}), \ \  x_{1} < i < x_{2}.
\end{array}
\right.
\end{equation}

Here, $x_1$ and $x_2$ are the location of the edge of the line-segment boundary, which is to be optimized during inverse optimization. $GM_{-1}(|i - x_{1}|, |i - x_{2}|)=(\frac{1}{2}(|i - x_{1}|^{-1}+|i - x_{2}|^{-1}))^{-1}$ denotes the harmonic mean\footnote{$GM_{\gamma}(x,y)=(\frac{1}{2}(x^{\gamma}+y^{\gamma}))^{1/\gamma}$ is generalized mean with order $\gamma$. The harmonic mean $GM_{-1}(x, y)$ interpolates between arithmetic mean $GM_{1}(x, y)=\frac{1}{2}(x+y)$ and the minimum $GM_{-\infty}(x, y)=\text{min}(x,y)$, and is influenced more by the smaller of $x$ and $y$.}, which is influenced more by the smaller of $|i - x_{1}|$ and $|i - x_{2}|$, so it is a \emph{soft} version of the distance to the nearest edge inside the line segment of $x_1<i<x_2$. When $\beta$ tends to $0$, the function $f$ converges to a binary valued function: see also Fig. \ref{fig:interpolation}. %\tailin{Need also explain how to construct the function with multiple obstacles, etc.}

\makeatletter 
\renewcommand{\thefigure}{S\arabic{figure}}
\makeatother
\setcounter{figure}{1}
\begin{figure}[h]
\begin{center}
\includegraphics[width=0.6\columnwidth]{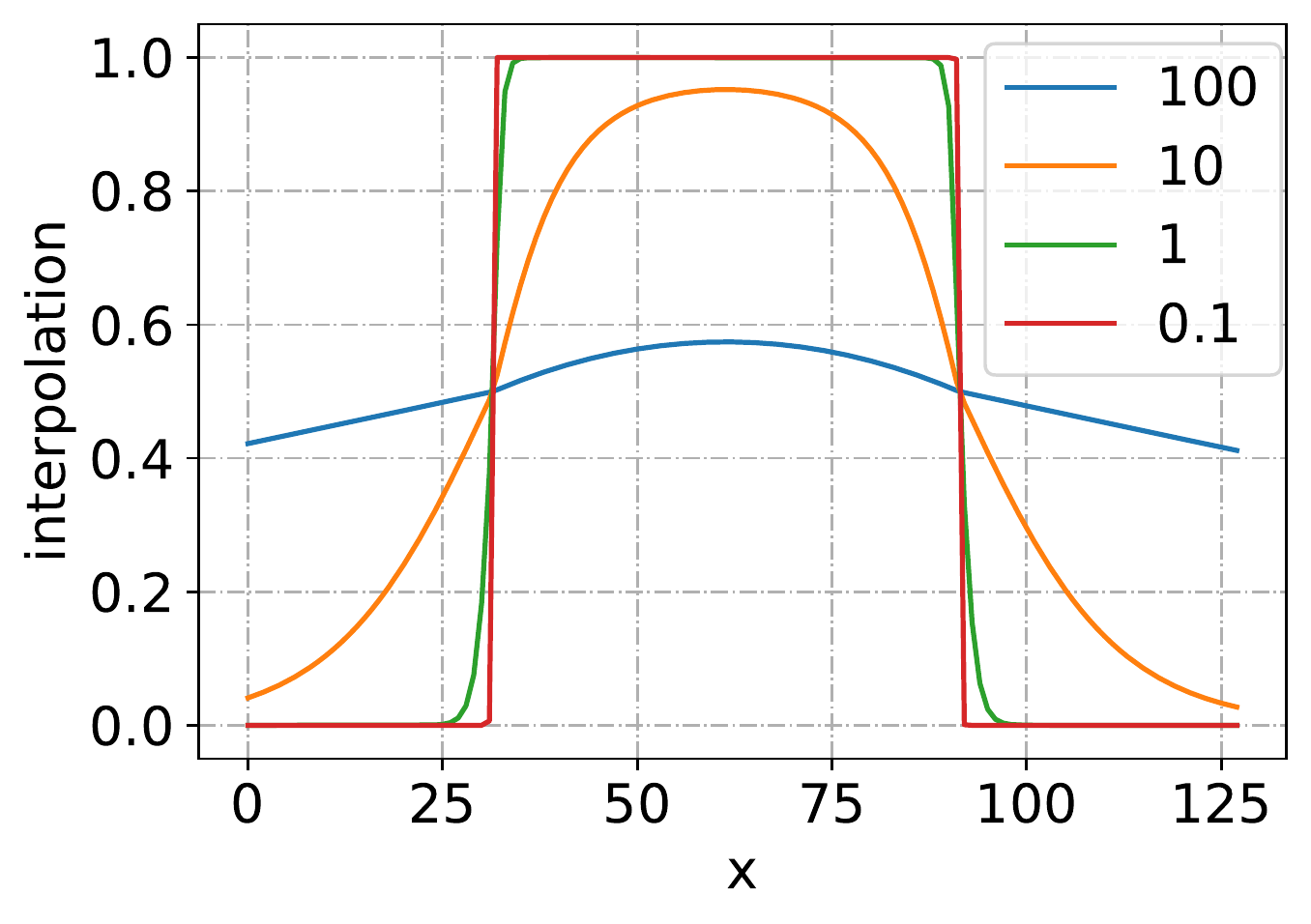}
\end{center}
\caption{The interpolation of binary valued function by a sigmoid-interpolation function. Continuous variables $(x_{1}, x_{2})$ are set to be $(31.5, 91.3)$. The continuous variables define edges of a continuous segment.}

\label{fig:interpolation}
\end{figure}
We define a continuous boundary function CB on a segment in a grid to be the pullback of the sigmoid-interpolation function with the projection to $1$-dimensional discretized line (\textit{i.e.}, take a projection of the pair of integers onto a $1$-dimensional segment and apply $f$):
\begin{equation}
\label{eq:cb}
CB(\ (i, j)\ |\ (x_{1}, x_{2}), \beta) =  \left\{
\begin{array}{ll}
f(\ i\ |\ (x_{1}, x_{2}), \beta), \ \ \ \  \text{if $(i, j)$ is in a horizontal segment}, \\
f(\ j\ |\ (x_{1}, x_{2}), \beta), \ \ \ \  \text{if $(i, j)$ is in a vertical segment.}
\end{array}
\right.
\end{equation}

\begin{figure}[t]
\begin{center}
\includegraphics[width=\columnwidth]{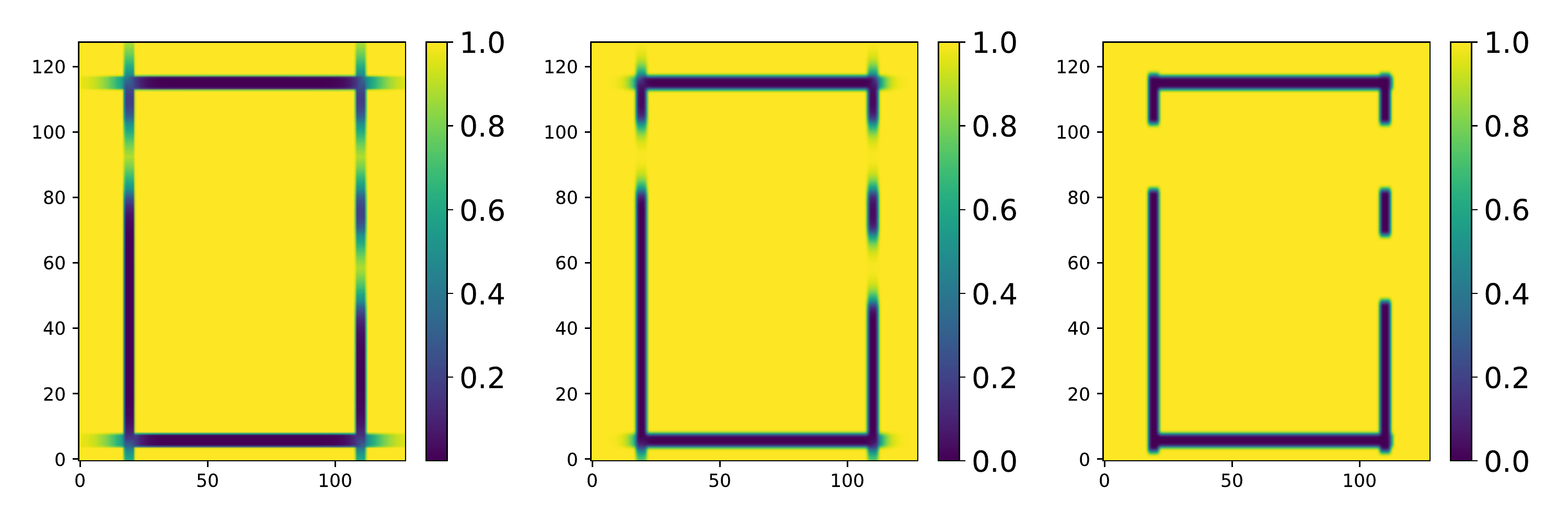}
\end{center}
\caption{Continuous bounds with different parameters $\beta =$ $5$ (left), $2$ (middle) and $0.01$ (right). As $\beta$ decreases, the edges of the boundaries tend to have steeper slopes.}
\label{fig:conti_bounds}
\end{figure}

Finally, a continuous boundary mask on a grid is obtained by (tranformation by a function $1-x$ and) taking the maximum on a set of $CB$s on boundary segments on the grid (see also Fig. \ref{fig:conti_bounds}). The boundary interpolation allows the gradient to pass through the boundary mask and able to optimize the location of the edge of line segments (\textit{e.g.} $x_1, x_2$).

\xhdr{Boundary annealing} As we see above, $\beta$ can be seen as a temperature hyperparameter, and the smaller it is, the more the boundary mask approximates a binary valued mask, and the less cells the boundary directly influences. At the beginning of the optimization, the parameter of the boundary (locations $x_1,x_2$ of each line segment) may be far away from the optimal location. Having a small temperature $\beta$ would result in vanishing gradient for the optimization, and  very sparse interaction where the boundary mainly interact with its immediate neighbors, resulting in that very small gradient signal to optimize. Therefore, we introduce an annealing technique for the boundary optimization, where at the beginning, we start at a larger $\beta_0$, and linearly tune it down until at the end reaching a much smaller $\beta$. The larger $\beta$ at the beginning allows denser gradient at the beginning of inverse optimization, where the location of the boundary can also influence more cells, producing more gradient signals. The smaller $\beta$ at the end allows more accurate boundary location optimization at the end, where we want to reduce the bias introduced by the boundary interpolation.

\section{Model Architecture for \proj}
\label{app:model_architecture}
 
Here we detail the architecture of \proj, complementary to Sec. \ref{sec:model_architecture}. This architecture is used throughout all experiment, with just a few hyperparameter (\textit{e.g.} latent dimension $d_z$, number of convolution layers) depending on the dimension (1D, 2D, 3D) of the problem.  We first detail the 4 architectural components of \proj, and then discuss its current limitations.

\xhdr{Dynamic encoder $q$} The dynamic encoder $q$ consists of one CNN layer with (kernel-size, stride, padding) = $(3,1,1)$ and ELU activation, followed by $F_q$  convolution blocks, then followed by a flatten operation and an MLP with 1 layer and linear activation that outputs a $d_z$-dimensional vector $\z^k\in \R^{d_z}$ at time step $k$. Each of the $F_q$ convolution block consists of a convolution layer with (kernel-size, stride, padding) = $(4,2,1)$ followed by group normalization [74] (number of groups=2) and ELU activation [75]. The channel size of each convolution block follows the standard exponentially increasing pattern, \textit{i.e.} the first convolution block has $C$ channels, the second has $C\times 2^1$ channels, ... the $n^\text{th}$ convolution block has $C\times 2^{n-1}$ channels. The larger channel size partly compensates for smaller spatial dimensions of feature map for higher layers.

\xhdr{Static encoder $r$} For the static encoder $r$, depending on the static parameter $p$, it can be an $F_r$-layer MLP (as in 1D experiment Sec. \ref{sec:burgurs} and 3D experiment Appendix \ref{app:3d_nv}), or a similar CNN+MLP architecture as the dynamic encoder (as in Sec. \ref{sec:inverse_opt_exp} that takes as input the boundary mask). If using MLP, it uses $F_r$ layers with ELU activation and the last layer has linear activation. In our experiments, we select $F_r\in\{0,1,2\}$, and when $F_r=0$, it means no layer and the static parameter is directly used as $\z_p$. The static encoder outputs a $d_{zp}$-dimensional vector $\z_p\in \R^{d_{zp}}$.

\xhdr{Latent evolution model $g$} The latent evolution model $g$ takes as input the concatenation of $\z^k$ and $\z_p$ (concatenated along the feature dimension), and outputs the prediction  $\hat{\z}^{k+1}$. We model it as an MLP with residual connection from input to output, as an equivalent of the forward Euler's method in latent space:

\begin{equation}
\hat{\z}^{k+1}=\text{MLP}_g(\z)^k+\z^k
\end{equation}

In this work, we use the same $\text{MLP}_g$ architecture throughout all sections, where the $\text{MLP}_g$ consists of 5 layers, each layer has the same number $d_z$ of neurons as the dimension of $\z^k$. The first three layers has ELU activation, and the last two layers have linear activation. We use two layers of linear layer instead of one, to have an implicit rank-minimizing regularization [76], which we find performs better than 1 last linear layer.

\xhdr{Decoder $h$} Mirroring the encoder $q$, the decoder $h$ takes as input the $\z^{k+m}\in\R^{d_z},m=0,1,...M$, through an $\text{MLP}_h$ and a CNN with $F_h=F_q$ number of convolution-transpose blocks, and maps to the state $U^{k+m}$ at input space. The $\text{MLP}_h$ is a one layer MLP with linear activation. After it, the vector is reshaped into the shape of (batch-size, channel-size, *image-shape) for the $F_h$ convolution-transpose blocks. Then it is followed by a single convolution-transpose layer with (kernel-size, stride, padding)=$(3,1,1)$ and linear activation. Each convolution-transpose block consists of one convolution-transpose layer with (kernel-size, stride, padding) = $(4,2,1)$, followed by group normalization and an ELU activation. The number of channels also follows a mirroring of the encoder $q$, where the nearer to the output, the smaller the channel size with exponentially decreasing size.

\xhdr{Limitations of current \proj architecture} The use of MLPs in the encoder and decoder has its benefits and downside. The benefit is that due to its flatten operation and MLP that maps to a much smaller vector $\z$, it can significantly improve speed, as demonstrated in the experiments in the paper. The limitation is that it requires that the training and test datasets to have the same discretization, otherwise a different discretization will result in a different flattened dimension making the MLP in the encoder and decoder invalid. We note that despite this  limitation, it already encompasses a vast majority of applications where the training and test datasets share the same discretization (but with novel initial condition, static parameter $p$, etc.). Experiments in this paper show that our method is able to generalize to novel equations in the same family (Sec. \ref{sec:burgurs}), novel initial conditions (Sec. \ref{sec:2d_nv} and \ref{sec:inverse_opt_exp}) and novel Reynolds numbers in 3D (Appendix \ref{app:3d_nv}). Furthermore, our general 4-component architecture of dynamic encoder, static encoder, latent evolution model and decoder is very general and can allow future work to transcend this limitation. Future work may go beyond the limitation of discretization, by incorporating ideas from \textit{e.g.} neural operators \cite{li2020neural,lulearning}, where the latent vector encodes the solution \emph{function} $\u(\x,t)$ instead of the discretized states $U^k$, and the latent evolution model then models the latent dynamics of neural \emph{operators} instead of functions.

Similar to a majority of other deep-learning based models for surrogate modeling (\textit{e.g.} \cite{ sanchez2020learning,li2021fourier}), the conservation laws present in the PDE is \emph{encouraged} through the loss w.r.t. the ground-truth, but not generally \emph{enforced}. Building domain-specific architectures that enforces certain conservation laws is out-of-scope of this work, since we aim to introduce a more general method for accelerating simulating and inverse optimizing PDEs, applicable to a wide scope of temporal PDEs. It is an exciting open problem, to build more structures into the latent evolution that obeys certain conservation laws or symmetries, potentially incorporating techniques \textit{e.g.} in [77, 78]. Certain conservation laws can also be enforced in the decoder, for example similar to the zero-divergence as in \cite{kim2019deep}.

\section{Details for experiments in 1D family of nonlinear PDEs}
\label{app:1d_exp}

Here we provide more details for the experiment for Sec. \ref{sec:burgurs}. The details of the dataset have already been given in Section \ref{sec:burgurs} and more detailed information can be found in \cite{brandstetter2022message} that introduced the benchmark. 

\xhdr{\proj} For \proj in this section, the convolution and convolution-transpose layers are 1D convolutions, since the domain is 1D. We use temporal bundling steps $S=25$, similar to the MP-PDE, so it based on the past $S=25$ steps to predict the next $S=25$ steps. The input has shape of (batch-size, $S$, $C_\text{in}=1$, $n_x$) , which\footnote{Here $C_\text{in}$ is the number of input channels for $u(t,x)$. It is 1 since the $u(t,x)$ has only one feature.} we flatten the $S$ and $C_\text{in}$ dimensions into a single dimension and feed the (batch-size, $S\times C_\text{in}=25$, $n_x$) tensor to the encoder. For the convolution layers in encoder, we use starting channel size $C=32$ and exponential increasing channels as detailed in Appendix \ref{app:model_architecture}. We use $F_q=F_r=4$ blocks of convolution (or convolution-transpose). 

We perform search on hyperparameters of latent dimension $d_z\in\{64,128\}$, loss function $\ell\in\{\text{MSE},\text{RMSE}\}$, time horizon $M\in\{4,5\}$, and number of layers for static encoder $F_r\in\{0,1,2\}$, and use the model with the best validation loss. We train for 50 epochs with Adam \cite{kingma2014adam} optimizer with learning rate of $10^{-3}$ and cosine learning rate annealing [76] whose learning rate follows a cosine curve from $10^{-3}$ to $0$.

\xhdr{Baselines} For baselines, we directly report the baselines of MP-PDE, FNO-RNN, FNO-PR and WENO5 as provided in \cite{brandstetter2022message}. Details for the baselines is summarized in Sec. \ref{sec:burgurs} and more in \cite{brandstetter2022message}.

\xhdr{More explanation for Table \ref{tab:1d}} The runtimes in Table \ref{tab:1d} are for one full unrolling that predicts the future 200 steps starting at step 50, on a NVIDIA  2080 Ti RTX GPU. The ``full'' runtime includes the time for encoder, latent evolution, and decoding to all the intermediate time steps. The ``evo'' runtime only includes the runtime for the encoder and the latent evolution. The representation dimension, as explained in Sec. \ref{sec:burgurs}, is the number of feature dimensions to update at each time step. For baselines of MP-PDE, etc. it needs to update $n_x\times S\times1$ dimensions, \textit{i.e.} the consecutive $S=25$ steps of the 1D space with $n_x$ cells (where each cell have one feature). For example, for $n_x=100$, the representation dimension is $n_x\times S\times1=100\times 25\times1=2500$. In contrast, our \proj uses a 64 or 128-dimensional latent vector to represent the same state, and only need to update it for every latent evolution.

\xhdr{Visualization of \proj rollout} In Fig. \ref{fig:1d_rollout}, we show example rollout of our \proj in the \textbf{E2} scenario and comparing with ground-truth. We see that \proj captures the shock formation (around $x=14$) faithfully, across all three spatial discretizations.

\makeatletter 
\renewcommand{\thefigure}{\arabic{figure}}
\makeatother
\setcounter{figure}{3}
\begin{figure}[t]
\centering
\begin{subfigure}{0.49\textwidth}
    \includegraphics[width=\textwidth]{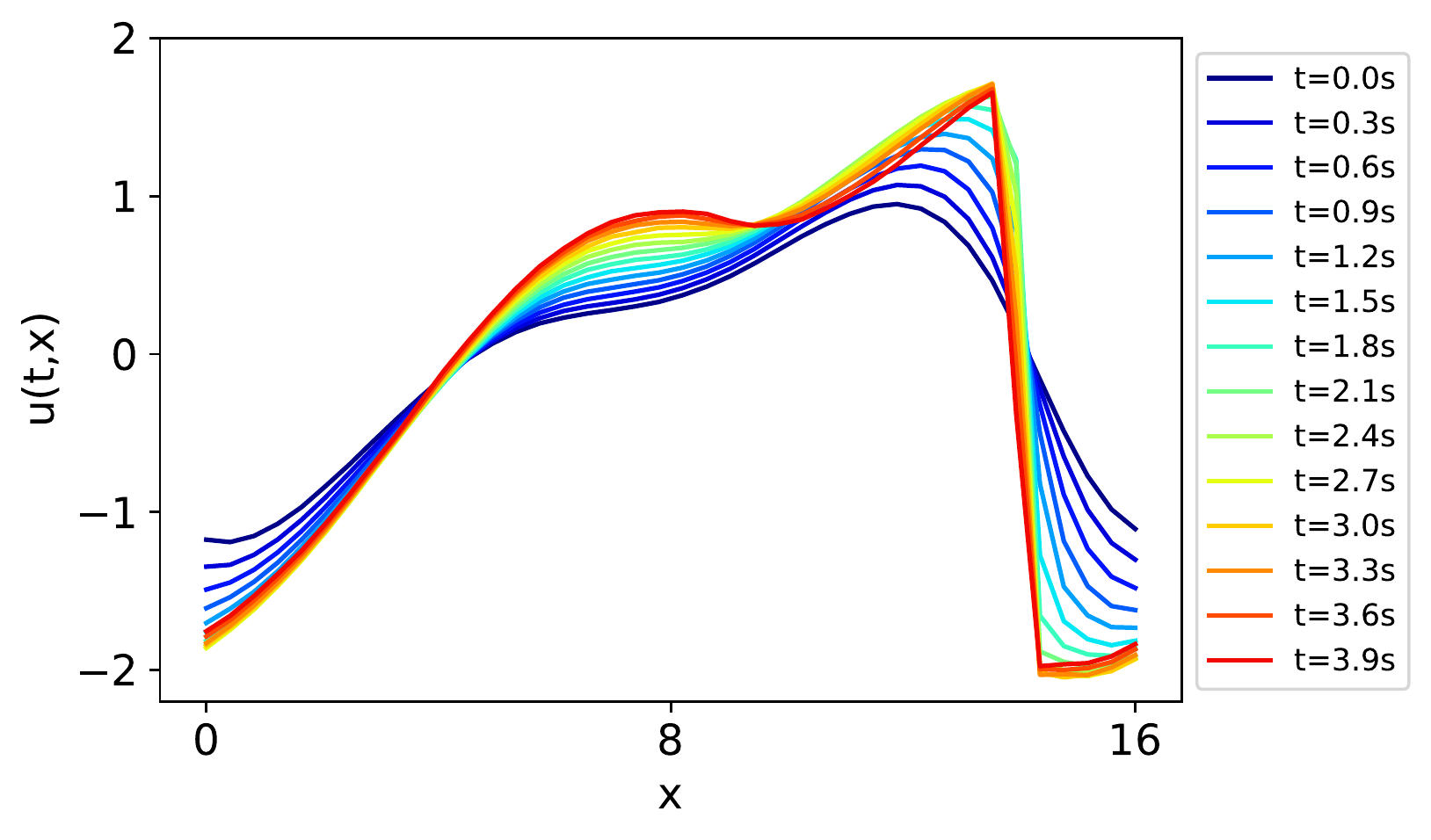}
    \caption{\proj  prediction with $n_x=40$}
    \label{fig:1d_E2_40}
\end{subfigure}
\hfill
\begin{subfigure}{0.49\textwidth}
    \includegraphics[width=\textwidth]{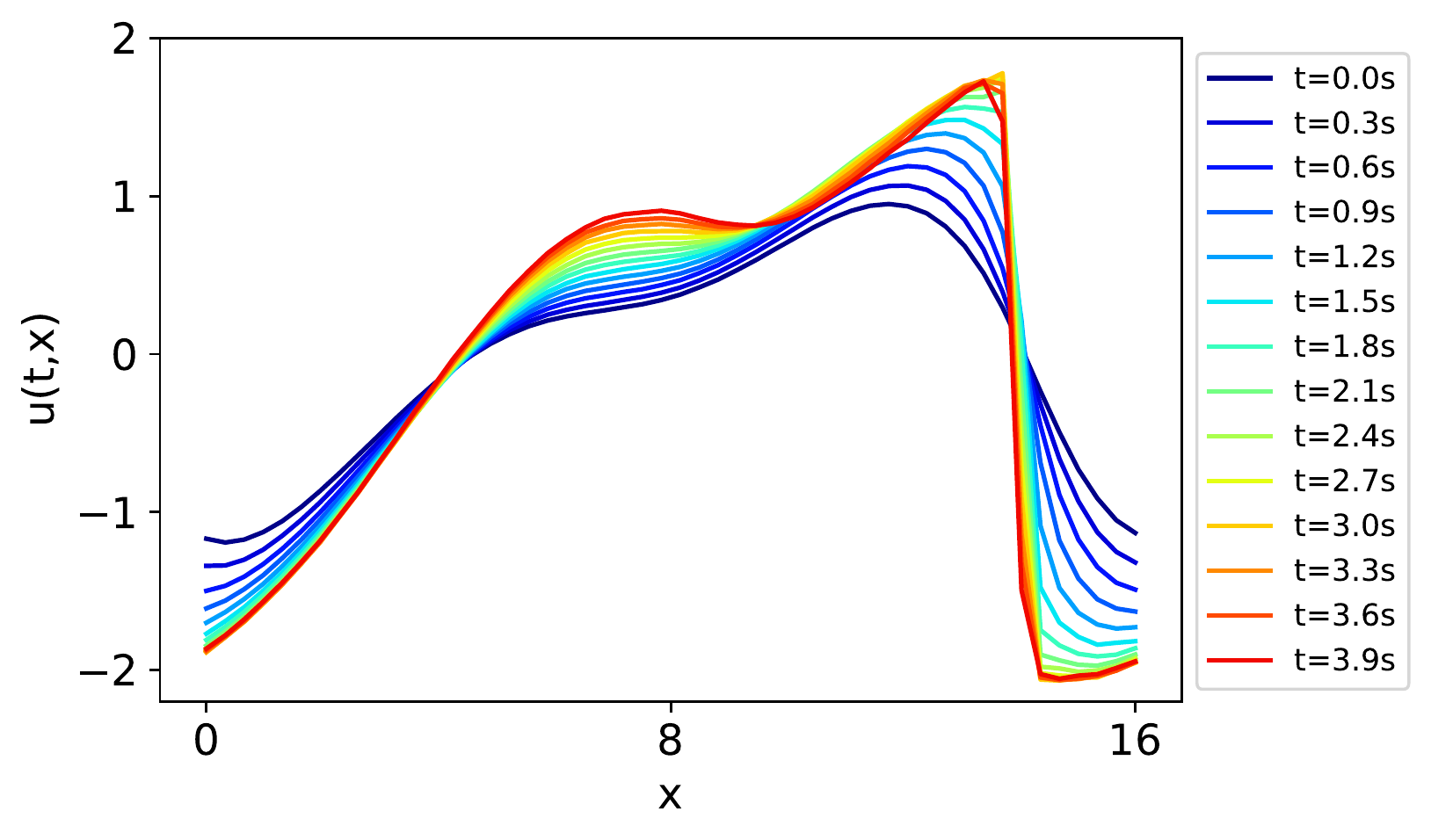}
    \caption{\proj prediction with $n_x=50$ }
    \label{fig:1d_E2_50}
\end{subfigure}
\begin{subfigure}{0.49\textwidth}
    \includegraphics[width=\textwidth]{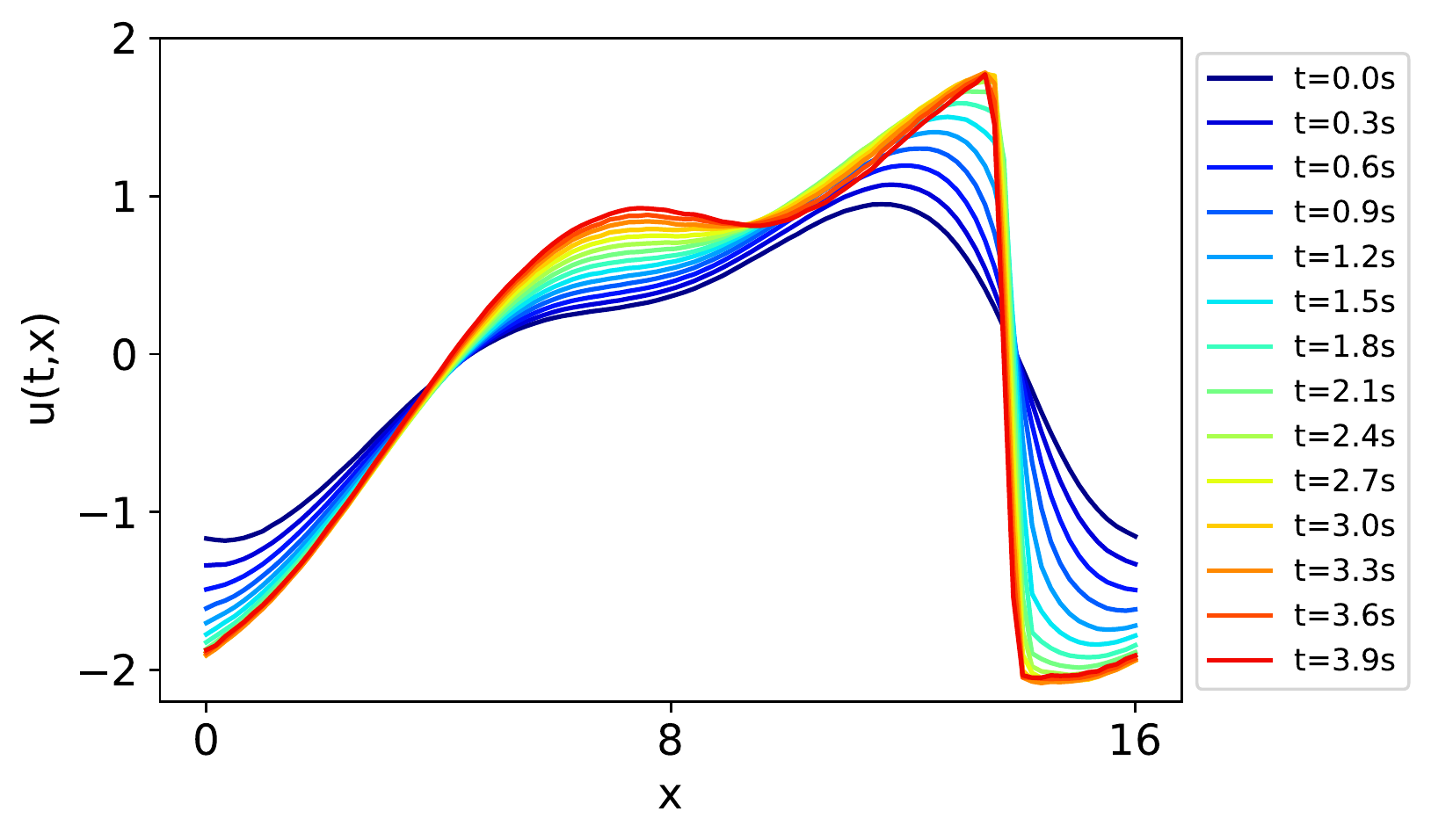}
    \caption{\proj  prediction with $n_x=100$}
    \label{fig:1d_E2_100}
\end{subfigure}
\hfill
\begin{subfigure}{0.49\textwidth}
    \includegraphics[width=\textwidth]{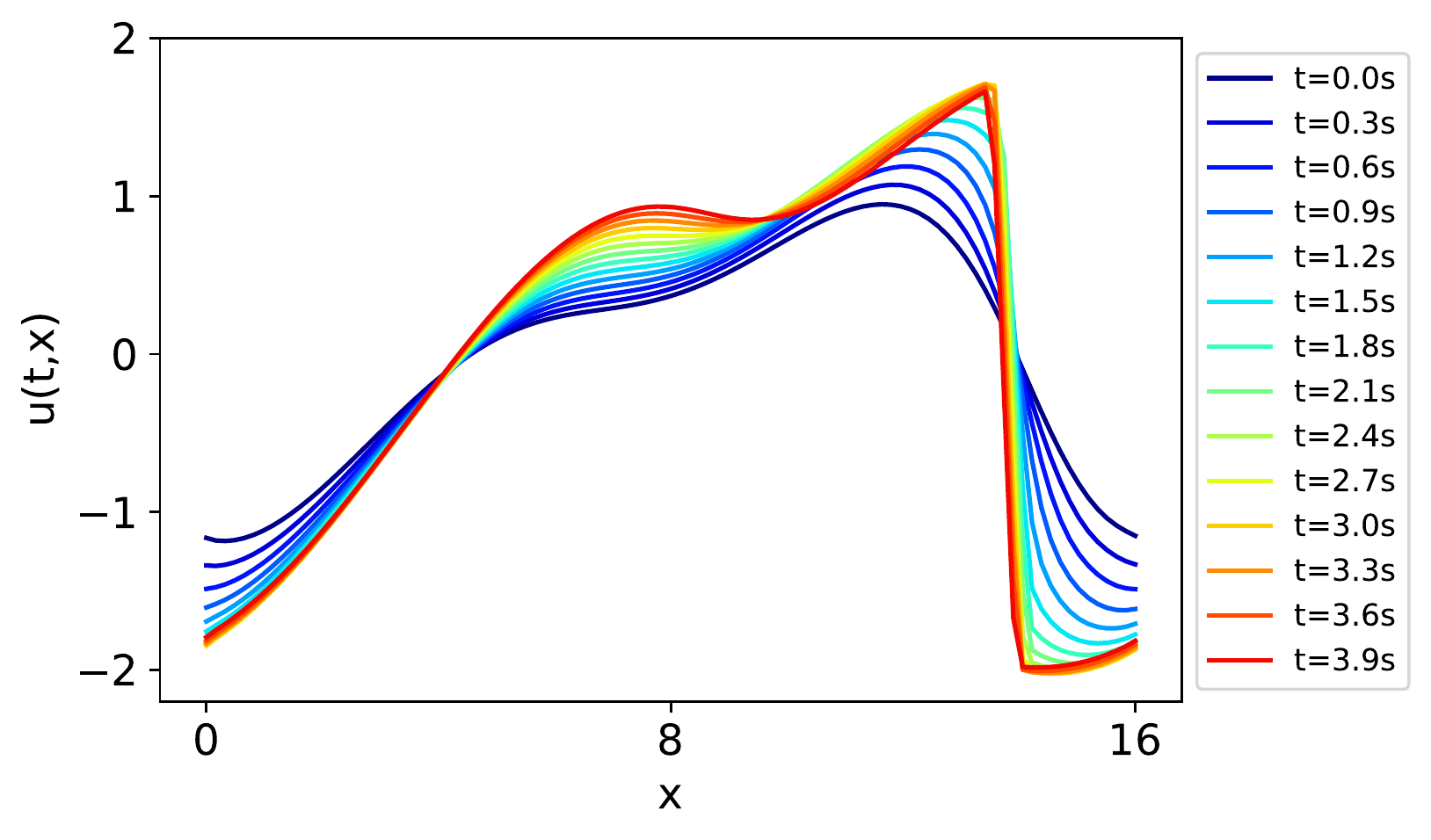}
    \caption{Ground-truth with $n_x=100$}
    \label{fig:1d_E2_gt}
\end{subfigure}
\caption{Example rollout of \proj for 200 steps (0 to 4s), with \textbf{E2} scenario that tests the models ability to generalize to new equations within the same family, for (a) $n_x=40$, (b) $n_x=50$, (c) $n_x=100$, compared with ground-truth of (d) $n_x=100$. The \proj models in the plot are using the ones reported in Table \ref{tab:1d}. We see that \proj captures the shock formation (around $x=14$) very accurately and faithfully, across all three spatial discretizations.}
\label{fig:1d_rollout}
\end{figure}

\section{Details for 2D Navier-Stokes flow}
\label{app:2d_nv}

Here we detail the experiments we perform for Sec. \ref{sec:2d_nv}. For the baselines, we use the results reported in \cite{li2021fourier}. For our \proj, we follow the same architecture as detailed in Appendix \ref{app:model_architecture}. Similar to other models (\textit{e.g.} FNO-2d), we use temporal bundling of $S=1$ (no bundling) and use the past 10 steps to predict one future step, and autoregressively rollout for $T-10$ steps, then use the relative L2 loss over the all the predicted states as the evaluation metric. We perform search on hyperparameters of latent dimension $d_z\in\{128,256\}$, loss function $\ell\in\{\text{MSE},\text{RMSE},\text{L2}\}$, time horizon $M\in\{4,T-10\}$, number of epochs $\{200,500\}$, and use the model with the best validation loss. The runtime in Table \ref{tab:2d_flow} is computed using an Nvidia Quadro RTX 8000 48GB GPU (since the FNO-3D exceeds the memory of the Nvidia 2080 Ti RTX 11GB GPU, to make a fair comparison, we use this larger-memory GPU for all models for runtime comparison).

\section{3D Navier-Stokes flow}
\label{app:3d_nv}
To explore how \proj can scale to larger scale turbulent dynamics and its potential speed-up, we train \proj in a 3D Navier-Stokes flow through the cylinder using a similar 3D dataset in \cite{um2020solver}, generated by PhiFlow \cite{phiflow} as the ground-truth solver. The PDE is given by:

\begin{align}
\partial_t u_x + \mathbf{u} \cdot \mathbf{\nabla} u_{x}  = - \frac{1}{\rho} \mathbf{\nabla} p + \nu \mathbf{\nabla} \cdot \mathbf{\nabla} u_{x}, &\\
\partial_t u_y + \mathbf{u} \cdot \mathbf{\nabla} u_{y}  = - \frac{1}{\rho} \mathbf{\nabla} p + \nu \mathbf{\nabla} \cdot \mathbf{\nabla} u_{y}, &\\
\partial_t u_ z+ \mathbf{u} \cdot \mathbf{\nabla} u_{z}  = - \frac{1}{\rho} \mathbf{\nabla} p + \nu \mathbf{\nabla} \cdot \mathbf{\nabla} u_{z}, &\\
\rm{subject\ to\ } \mathbf{\nabla} \cdot \mathbf{u} = 0.&
\end{align}

We discretize the space into a 3D grid of $256\times128\times128$, resulting in 4.19 million cells per time step. We generate $5$ trajectories of length $500$ with Reynolds number $\{55.5, 56.8, 58.0, 58.3, 58.6\}$ for training/validation set and test the model's performance on $2$ additional trajectories with $\{57.4, 58.0\}$. All the trajectories have different initial conditions. We sub-sample the time every other step, so the time interval between consecutive time step for training is 2s. For \proj, we follow the architecture in Appendix \ref{app:model_architecture}, with $F_q=F_h=5$ convolution (convolution-transpose) blocks in the encoder (decoder), latent dimension  $d_z=128$, and starting channel dimension of $C=32$. We use time horizon $M=4$ in the learning objective (Eq. \ref{eq:objective}), with $(\alpha_1,\alpha_2,\alpha_3,\alpha_4)=(1,0.1,0,0.1)$ (we set the third step $\alpha_3=0$ due to the limitation in GPU memory). The Reynolds number $p=Re$ is copied 4 times and directly serve as the static latent parameter (number of layers $F_r$ for static encoder MLP $r$ is 0). This static encoder allows \proj to generalize to novel Reynolds numbers. We use $\ell=$MSE. We randomly split 9:1 in the training/validation dataset of 5 trajectories, train for 30 epochs, save the model after each epoch, and use the model with the best validation loss for testing. 

\xhdr{Prediction quality} In Fig. \ref{fig:3d_nv}, we show the prediction of \proj on the first test trajectory with a novel Reynolds number ($Re=57.4$) and novel initial conditions. We see that \proj captures the high-level and low-level turbulent dynamics in a qualitatively reasonable way, both at the tail and also in the inner bulk. This shows the scalability of our \proj to learn large-scale PDEs with intensive dynamics in a reasonably faithful way.

\begin{table}[b]
\caption{Comparison of \proj with baseline on runtime and representation dimension, in the 3D Navier-Stokes flow. The runtime is to predict the state at $t=40$.}
\centering
\resizebox{1\textwidth}{!}{
\begin{tabular}{cccccccc}
\hline
& \makecell{Runtime (s)} & \makecell{Representation \\ dimension } & \makecell{Error at \\ $t=40$}& \# Paramters & \makecell{\# Parameters for \\ evolution model} & \makecell{Training time \\ (min) per epoch} & \makecell{Memory \\ usage (MiB)}\\
\hline
\makecell{PhiFlow (ground-truth \\ solver) on CPU} & 1802  & $16.76\times 10^6$  & - & - & - & - & -\\ 
\makecell{PhiFlow (ground-truth \\ solver) on GPU} & 70.80  & $16.76\times 10^6$ & - & - & - & - & -\\ 
FNO (with 2-step loss)& 7.00 & $16.76\times 10^6$ & \textbf{0.1695} & 3,281,864 & 3,281,864 & 102 & 25,147\\
FNO (with 1-step loss)& 7.00 & $16.76\times 10^6$ & 0.3215 & 3,281,864 & 3,281,864 & 58 & 24,891\\
\projnolatent  & 1.03 & $16.76\times 10^6$ & 0.1870 & 71,396,976 & 71,396,976 & 69 & 21,361\\
\hline
\textbf{\proj (ours)} & \textbf{0.084}  & \textbf{128} & 0.1947 & 65,003,120 & 83,072 & 65 & 25,595\\
\hline
\end{tabular}
}
\label{table:speed_3d}
\end{table}

\xhdr{Speed comparison} We compare the runtime of our \proj, an ablation \projnolatent and the ground-truth solver PhiFlow, to predict the state at $t=40$. The result is shown in Table \ref{table:speed_3d}. For the ablation \projnolatent, its  latent evolution model and the MLPs in the encoder and decoder are ablated, and it directly uses the other parts of encoder and decoder to predict the next step (essentially a 12-layer CNN). 
We see that our \proj achieves a $70.80/0.084\simeq 840\times$ speed up compared to the ground-truth solver on the same GPU. 
We see that w.r.t. \projnolatent (a CNN) that is significantly faster than solver, our \proj is still $1.03/0.084=12.3$ times faster. This shows that our \proj can significantly accelerate the simulation of large-scale PDEs.

\xhdr{Comparison of number of parameters} We see that our \proj uses much less number of parameters to evolve autoregressively than FNO. The most parameters of LE-PDE are mainly in the encoder and decoder, which is only applied once at the beginning and end of the evolution. Thus, \proj achieves a much smaller runtime than FNO to evolve to t=40.

\section{Details for inverse optimization of boundary conditions}
\label{app:inverse}

\makeatletter 
\renewcommand{\thefigure}{S\arabic{figure}}
\makeatother
\setcounter{figure}{3}
\begin{figure}[h]
\centerline{\includegraphics[scale=.31]{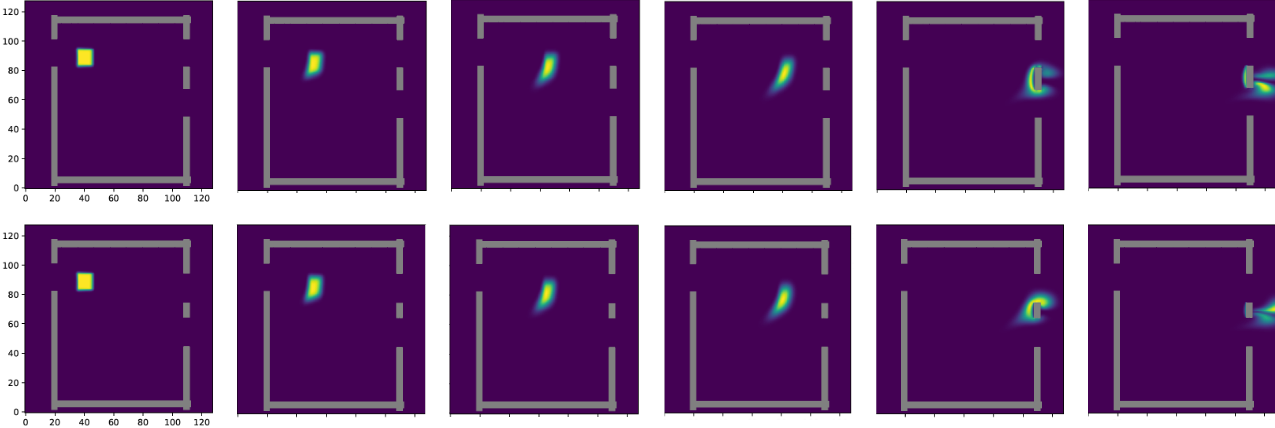}}
\caption{Trajectories generated by ground-truth solver with initial boundary parameter (upper) and optimized boundary parameter (lower).}
\label{fig:detail_inverse}
\end{figure}

\begin{figure}[h]
\begin{center}
\includegraphics[width=\columnwidth]{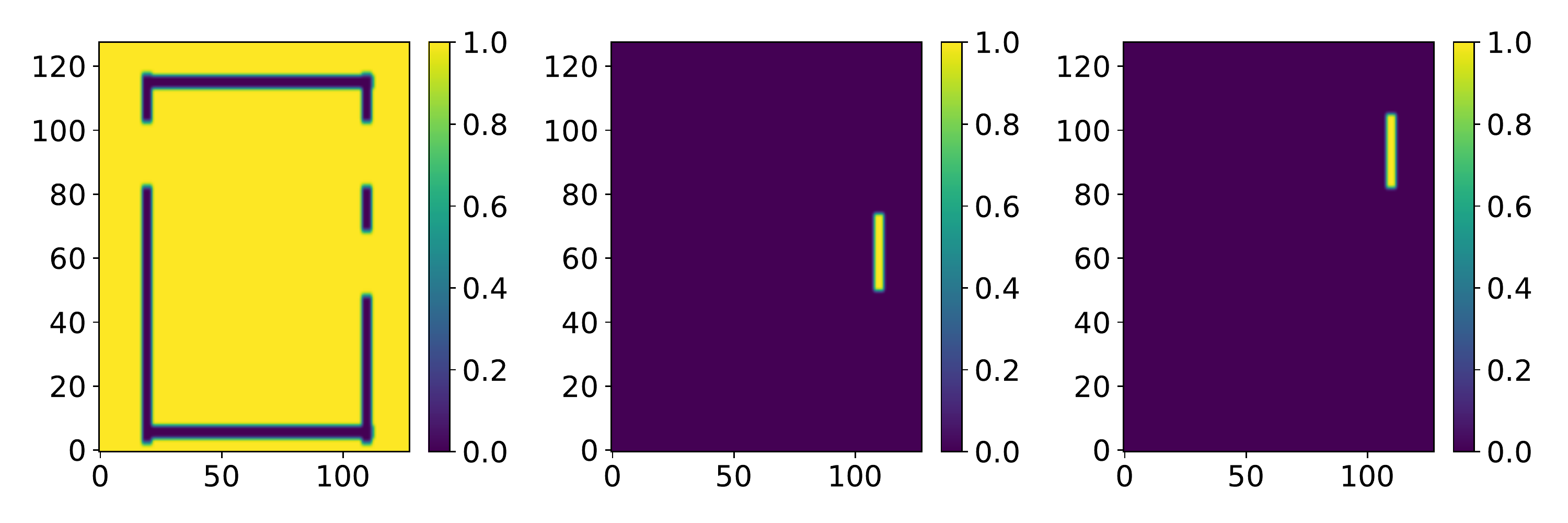}
\end{center}
\caption{Figures of outlet masks for given a boundary mask. The left mask is a boundary mask, the middle mask $o_{1}$ corresponds to the lower outlet and the right $o_{2}$ the upper outlet.}
\label{fig:outlet_masks}
\end{figure}

\xhdr{Objective function} To define the objective function, we create masks $(o_{1}, o_{2})$ that correspond to respective outlets of given a boundary. The masks are defined to be ones on the outlets' voids  (see also Fig. \ref{fig:outlet_masks}).
With the masks, we define the objective function in Sec. \ref{sec:inverse_optimization} that can measure the amount of smoke passing through the outlets:
\[ L_d[p] = \sum_{i=1}^{2} \operatorname{MSE}(t_{i}, \frac{\sum_{m=k_{s}}^{k_{e}} \langle o_{i}, \hat{U}^m(p) \rangle}{K}).\]
Here, $(t_{1}, t_{2}) = (0.3, 0.7)$, $K = \sum_{j=1}^{2}\sum_{m=k_{s}}^{k_{e}} \langle o_{j}, \hat{U}^m(p) \rangle$ and $ \langle x, y \rangle = x^{\operatorname{T}} y$. We set $k_{s} = 50$, \textit{i.e.},  we use smoke at scenes after $50$ time steps to calculate the amount of the smoke. 

\xhdr{\proj} The encoder $q$ and decoder $h$ have $F_{q} = F_{h} = 4$ blocks of convolution (or convolution-transpose) followed by MLP, as specified in Appendix \ref{app:model_architecture}. The time step of input is set to be $1$. The output of $q$ is a $128$-dimensional vector $\z^{k}$. The latent evolution model $g$ takes as input the concatenation of $\z^{k}$ and $16$-dimensional latent boundary representation $\z_{p}$ along the feature dimension, and outputs the prediction of $\hat{\z}^{k+1}$. Here, $\z_{p}$ is transformed by $r$ with the same layers as $q$, taking as input an boundary mask, where the boundary mask is a interpolated one specified in Appendix \ref{app:boundary_interpolation}. The architecture of the latent evolution model $g$ is the same as stated in Appendix \ref{app:model_architecture}, with latent dimension $d_z=128$.

\begin{figure}[h]
\centering
\begin{subfigure}{0.49\textwidth}
    %\includegraphics[width=\textwidth]{figures/init_traj.png}
    %\caption{Trajectory generated by ground-truth solver with an initial randomly generated boundary parameter.}
    %\label{fig:first}
    \centerline{\includegraphics[scale=.5]{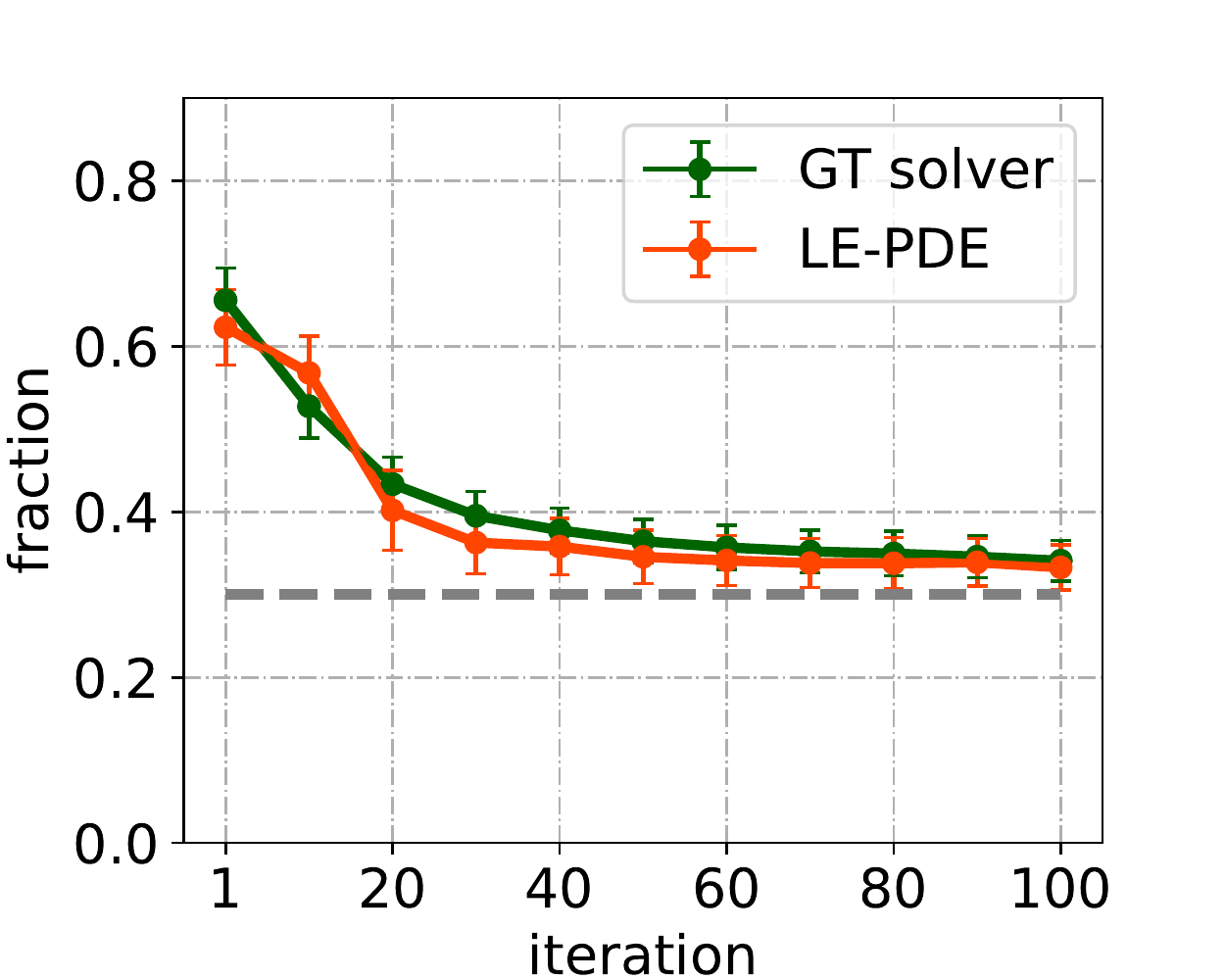}}
    \caption{Transition of fraction estimated by \proj with fixed $\beta$. The difference $(0.009)$ from fraction estimated by GT-solver is larger than that of \proj with annealing $(0.001)$ in Table \ref{table:invcomparison}.}
    \label{fig:ablation_fraction_gt_lepde}
\end{subfigure}
\begin{subfigure}{0.49\textwidth}
    %\resizebox{0.5\textwidth}{!}{
    \begin{tabular}{cc}
    \hline
    \textbf{\proj (ours)} & \makecell{GT-solver Error\\ (Model estimated Error)} \\
    \hline
    constant $\beta$ & 0.041 (0.032)  \\ 
    linear annealing $\beta$ & \textbf{0.035} (0.036)  \\
    \hline
    \end{tabular}
    \caption{Fractions estimated by ablated version of the inverse optimizer. Continuous boundary parameter $\beta$ in the ablated version is fixed across the iteration.}
    %}
    \label{table:ablation_fraction_gt_lepde}
\end{subfigure}
\caption{Ablation study of annealer in the inverse design for continuous boundary parameter $\beta$.}
\label{fig:ablation_fraction}
\end{figure}

\xhdr{Parameters for inverse design} We randomly choose 50 configurations for initial parameters. The sampling space is defined by the product of sets of inlet locations $\{79, 80, 81\}$, lower outlet locations $\{44, 45, 46, 47, 48, 49, 50\}$ and smoke position $\{0, 1\} \times \{-1, 0, 1\}$. We note that, even though we use the integers for the initial parameters, we can also use continuous values as initial parameters as long as the values are within the ranges of the integers. For one initial parameter, the number of the iterations of the inverse optimization is 100.  During the iteration for each sampled parameter, we also perform linear annealing for $\beta$ of continuous boundary mask starting from $0.1$ to $0.05$. We also perform an ablation experiment with fixed $\beta=0.05$ across the iteration. Fig. \ref{fig:ablation_fraction} shows the result. We see that without annealing, the GT-solver (ground-truth solver) computed Error (0.041) is larger than with annealing (0.035), and the gap estimated by the model and the GT-solver is much larger. This shows the benefit of using boundary annealing.

\begin{figure}[t]
\centering
\begin{subfigure}{0.49\textwidth}
    \includegraphics[width=\textwidth]{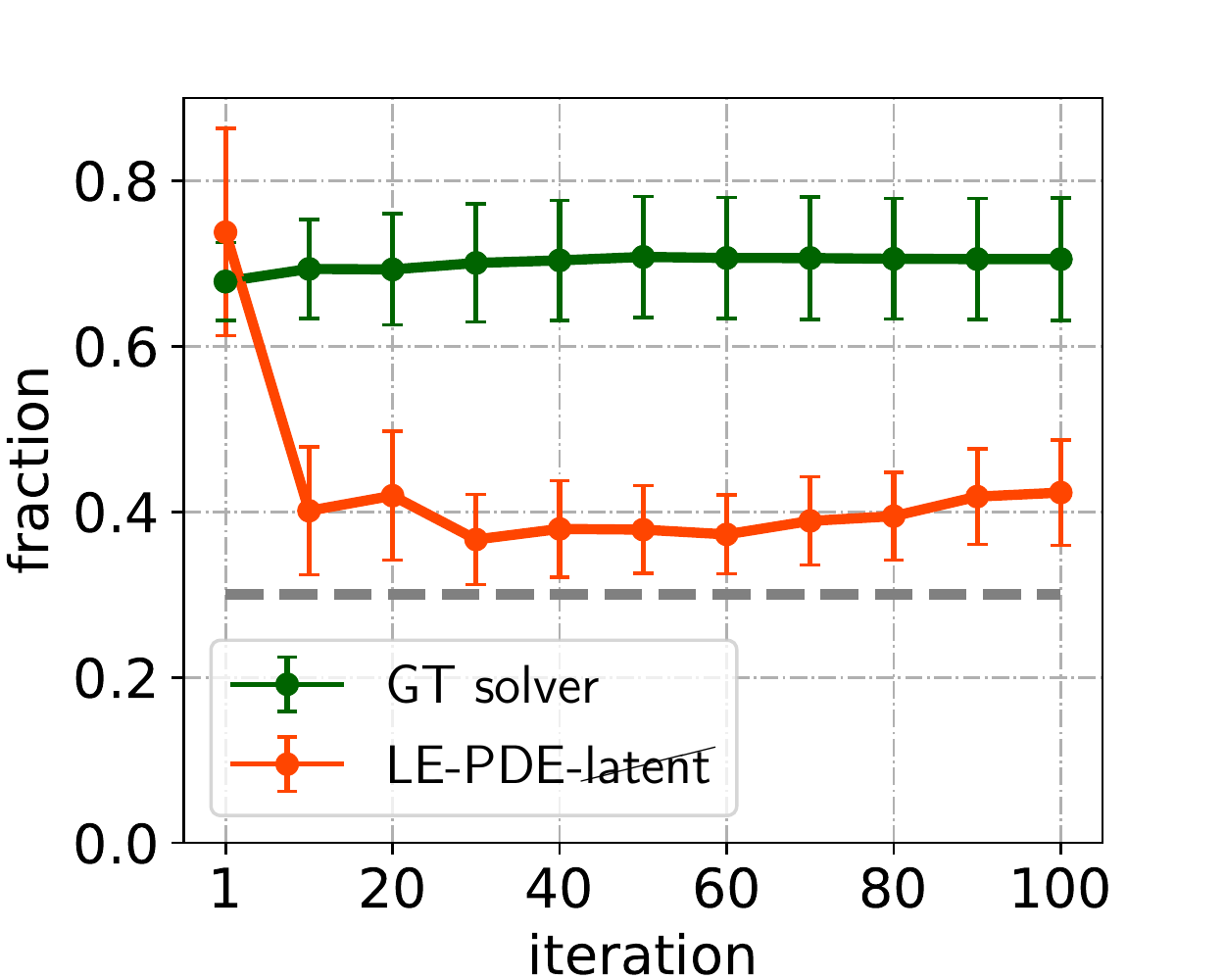}
    \caption{\projnolatent}
    \label{fig:fraction_nonlat}
\end{subfigure}
\begin{subfigure}{0.49\textwidth}
    \includegraphics[width=\textwidth]{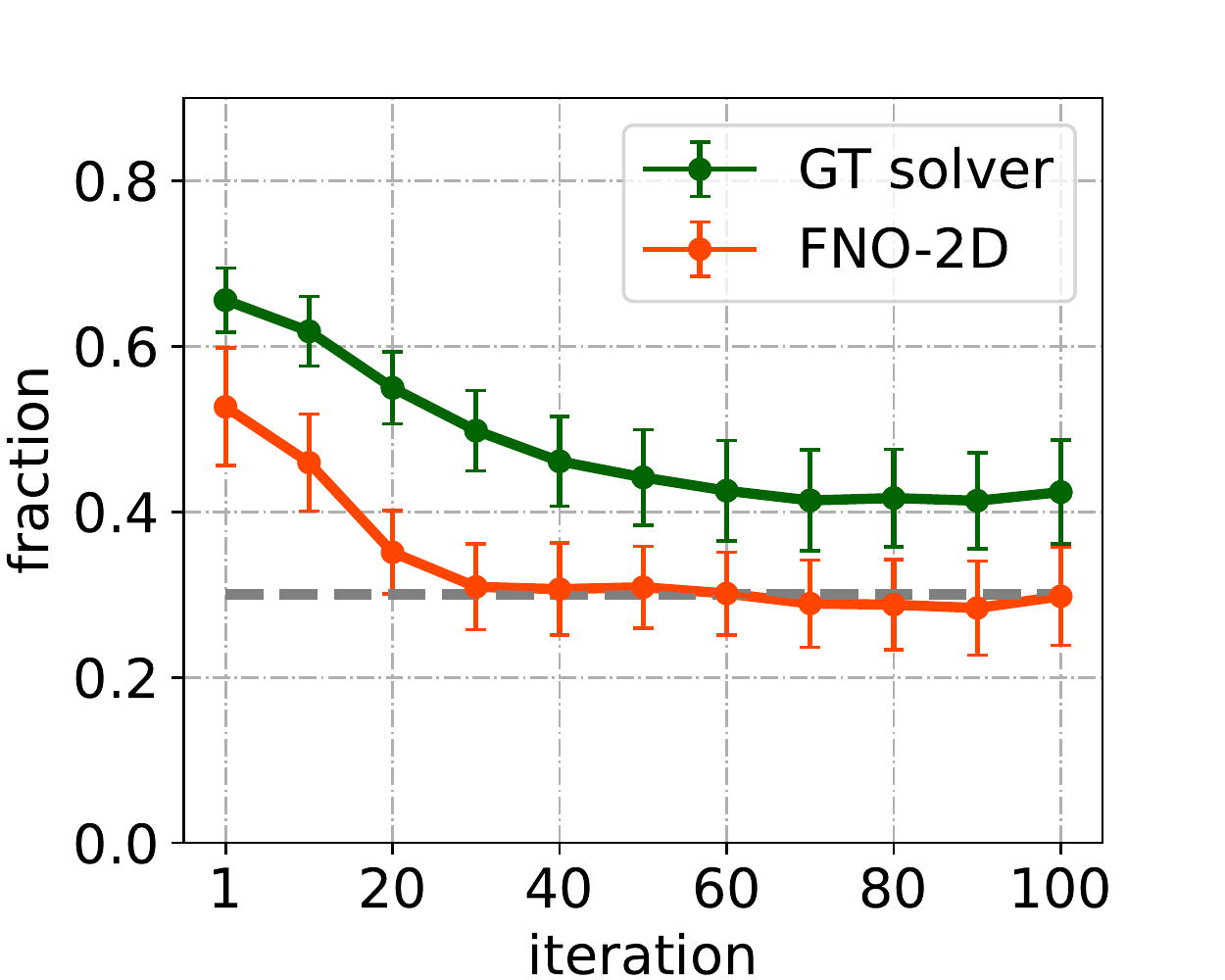}
    \caption{FNO-2D}
    \label{fig:fraction_fno}
\end{subfigure}
\caption{Fraction of smoke passing through the lower outlet computed by GT solver and estimated by \projnolatent and FNO-2D in Sec. \ref{sec:inverse_opt_exp}. The dashed line denotes the objective of 0.3 fraction of smoke passing through the lower outlet.}
\label{fig:fno_nonlat}
\end{figure}

\xhdr{Model architecture of baselines}
We use the same notation used in Appendix \ref{app:model_architecture}. \projnolatent uses the dynamic encoder $q$ subsequently followed by the decoder $h$. Both $q$ and $h$ have the same number of layers $F_{q} = F_{h} = 4$. The output of $h$ is used as the input of the next time step. For the FNO-2D model, we use the same architecture proposed in \cite{li2021fourier} with modes $= 12$ and width $= 20$. Fig. \ref{fig:fraction_nonlat} and \ref{fig:fraction_fno} are transition of fractions estimated by the ground-truth solver and the models with the boundary parameter under the inverse design. Compared with the one by our \proj in Fig. \ref{fig:forth}, we see that \proj has much better GT-solver estimated fraction, and less gap between the fraction estimated by the GT-solver and the model.

\section{More ablation experiments with varying latent dimension}
\label{app:ablation}

In this section, we provide complementary information to Sec. \ref{sec:ablation}. Specifically, we provide tables and figures to study  how the latent dimension $d_z$ influences the rollout error and runtime. Fig. \ref{fig:ablation} visualizes the results. 
Table \ref{table:ablation_latent_size_1d} shows the results in the 1D \textbf{E2} $(n_t,n_x)=(250,50)$ scenario that evaluate how \proj is able to generalize to novel PDEs within the same family. And Table \ref{table:ablation_latent_size_2d} shows the results in the 2D most difficult $(\nu=10^{-5},N=1000)$ scenario.

\xhdr{1D dataset} From Table \ref{table:ablation_latent_size_1d} and Fig. \ref{fig:ablation_1D_error}, we see that when latent dimension $d_z$ is between 16 and 128, the accumulated MSE is near the optimal of $1\sim 1.1$. It reaches minimum at $d_z=64$. With larger latent dimensions, \textit{e.g.} 256 or 512, the error slightly increases, likely due to the overfitting. With smaller latent dimension ($<8$), the accumulated error grows significantly. This shows that the intrinsic dimension of this 1D problem with temporal bundling of $S=25$ steps, is somewhere between 4 and 8. Below this intrinsic dimension, the model severely underfits, resulting in huge rollout error.

From the ``runtime full'' and ``runtime evo'' columns of Table \ref{table:ablation_latent_size_1d} and also in Fig. \ref{fig:ablation_1D_runtime}, we see that as the latent dimension $d_z$ decreases down from 512, the ``runtime evo'' has a slight decreasing trend down to 256, and then remains relatively flat. The ``runtime full'' also remains relatively flat. We don't see a significant decrease in runtime with decreasing $d_z$, likely due to that the runtime does not differ much in GPU with very small matrix multiplications.
\begin{table}[h]
\centering
\caption{Performance evaluation for \proj with different latent dimension on 1D dataset (E2-50 scenario). The accumulated error $=$ $\frac{1}{n_x}\sum_{t,x}\text{MSE}$, summing over the predicted steps of 50-250, the same as in Table \ref{tab:1d}. The runtime is measured by rolling out with the same 200 steps, measured on a NVIDIA 2080 Ti RTX GPU, same as in Table \ref{tab:1d}. The default is with $d_z=128$.)}
\begin{tabular}{l|c|c|c|c|c}
\hline
\makecell{\textbf{\proj} \\ \textbf{setting}} & \makecell{\textbf{cumulative} \\ \textbf{error}} & \makecell{\textbf{runtime}\\ \textbf{(full) (ms)}}               & \makecell{\textbf{runtime}\\ \textbf{ (evolution)} \\ \textbf{(ms)}}           & \makecell{\textbf{\# parameters}}                   & \makecell{\textbf{\# parameters for} \\ \textbf{latent evolution}\\ \textbf{ model}} \\
\hline
$d_z=512$       & 2.778            &  16.3  $\pm$ 2.6 &  6.7  $\pm$ 1.0 &  4043648 &  1314816          \\
$d_z=256$       & 2.186            &  15.0  $\pm$ 0.8 &  6.1  $\pm$ 0.3 &  2271360 &  329728           \\
$d_z=128$       & 1.127            &  14.9  $\pm$ 1.1 &  6.0  $\pm$ 0.4 &  1630976 &  82944            \\
$d_z=64$        & 0.994            &  14.4  $\pm$ 1.0 &  5.7  $\pm$ 0.3 &  1372224 &  20992            \\
$d_z=32$        & 1.048            &  14.5  $\pm$ 0.8 &  5.8  $\pm$ 0.4 &  1258208 &  5376             \\
$d_z=16$        & 1.041            &  14.1  $\pm$ 0.9 &  5.8  $\pm$ 0.4 &  1205040 &  1408             \\
$d_z=8$         & 21.03            &  14.0  $\pm$ 0.7 &  5.6  $\pm$ 0.2 &  1179416 &  384              \\
$d_z=4$         & 205.09           &  13.9  $\pm$ 0.5 &  5.7  $\pm$ 0.3 &  1166844 &  112             \\
\hline
\end{tabular}
\label{table:ablation_latent_size_1d}
\end{table}

\xhdr{2D dataset} From Table \ref{table:ablation_latent_size_2d} and Fig. \ref{fig:ablation_2D_error}, we see that similar to the 1D case, the error has a minimum in intermediate values of $d_z$. Specifically, as the latent dimension $d_z$ decreases from 512 to 4, the error first goes down and reaching a minimum of 0.1861 at $d_z=128$. Then it slightly increase with decreasing $d_z$ until $d_z=16$. When $d_z<16$, the error goes up significantly. This shows that large latent dimension may results in overfitting, and the intrinsic dimension for this problem is somewhere between 8 and 16, below which the error will significantly go up. As the latent dimension decreases, the runtime have a very small amount of decreasing (from 512 to 256) but mostly remain at the same level. This relatively flat behavior is also likely due to that the runtime does not differ much in GPU with very small matrix multiplications.

\begin{table}[h]
\centering
\caption{Performance evaluation for \proj with different latent dimension on 2D dataset ($\nu=10^{-5}$ scenario. The Error is the relative L2 norm measured over 10 rollout steps, the same as in Table \ref{tab:2d_flow}. The runtime is measured by rolling out with the same 10 steps, measured on a Nvidia Quadro RTX 8000 48GB GPU (same as in Table \ref{tab:2d_flow}), and average over 100 runs (the number after $\pm$ is the std. of the 100 runs). The default is with $d_z=128$.)}
\begin{tabular}{l|c|c|c|c|c}
\hline
\makecell{\textbf{\proj}\\ \textbf{setting}} & \makecell{\textbf{cumulative}\\ \textbf{error}} & \makecell{\textbf{runtime}\\ \textbf{(full) (ms)}}                 & \makecell{\textbf{runtime}\\ \textbf{(evolution)}\\ \textbf{(ms)}}           & \makecell{\textbf{\# parameters}}                   & \makecell{\textbf{\# parameters for}\\ \textbf{latent evolution}\\ \textbf{model}}          \\
\hline
$d_z=512$       & 0.1930           &  16.2 $\pm$ 1.1 &  6.8 $\pm$ 0.7 &  6467184 &  1313280 \\
$d_z=256$       & 0.1861           &  14.8 $\pm$ 1.1 &  5.8 $\pm$ 0.4 &  3384944 &  328960  \\
$d_z=128$       & 0.2064           &  14.8 $\pm$ 0.5 &  5.9 $\pm$ 0.4 &  2089584 &  82560   \\
$d_z=64$        & 0.2252           &  14.7 $\pm$ 0.7 &  6.0 $\pm$ 0.7 &  1503344 &  20800   \\
$d_z=32$        & 0.2315           &  15.0 $\pm$ 2.1 &  5.9 $\pm$ 0.5 &  1225584 &  5280    \\
$d_z=16$        & 0.2236           &  14.2 $\pm$ 1.3 &  5.8 $\pm$ 0.6 &  1090544 &  1360    \\
$d_z=8$         & 0.3539           &  14.3 $\pm$ 0.6 &  5.7 $\pm$ 0.3 &  1023984 &  360     \\
$d_z=4$         & 0.6353           &  14.2 $\pm$ 0.5 &  5.7 $\pm$ 0.2 &  990944  &  100    \\
\hline
\end{tabular}
\label{table:ablation_latent_size_2d}
\end{table}

\makeatletter 
\renewcommand{\thefigure}{\arabic{figure}}
\makeatother
\setcounter{figure}{5}
\begin{figure}[th!]
\centering
\begin{subfigure}{0.44\textwidth}
    \includegraphics[width=\textwidth]{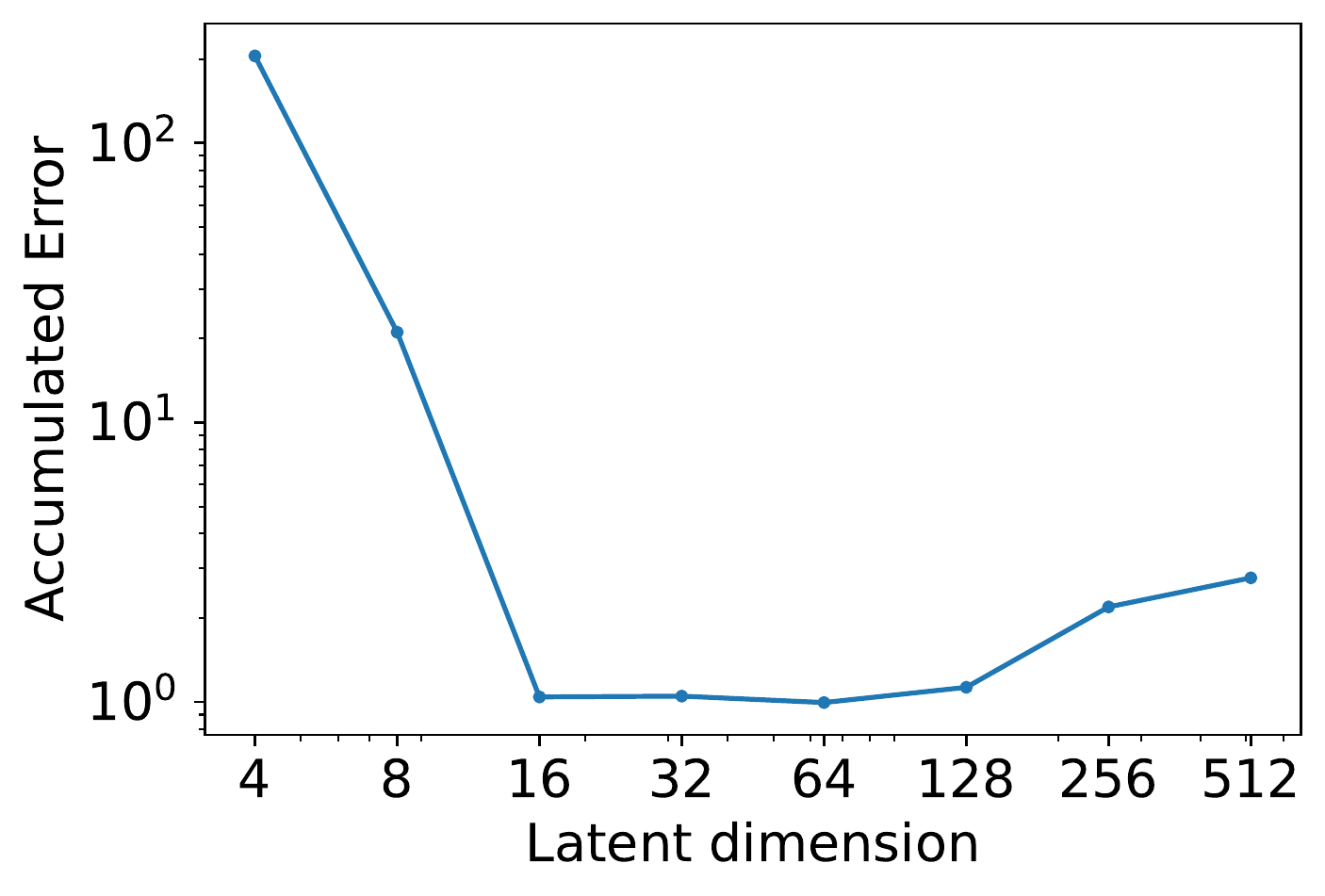}
    \caption{Accumulated Error vs. latent dimension  in 1D scenario. The Error is the relative L2 norm measured over 10 rollout steps, the same as in Table \ref{tab:2d_flow}. The runtime is measured by rolling out with the same 10 steps, measured on a Nvidia Quadro RTX 8000 48GB GPU (same as in Table \ref{tab:2d_flow}), and average over 100 runs (the number after $\pm$ is the std. of the 100 runs). The default is with $d_z=128$.}
    \label{fig:ablation_1D_error}
\end{subfigure}
\hfill\hfill
\begin{subfigure}{0.55\textwidth}
    \includegraphics[width=\textwidth]{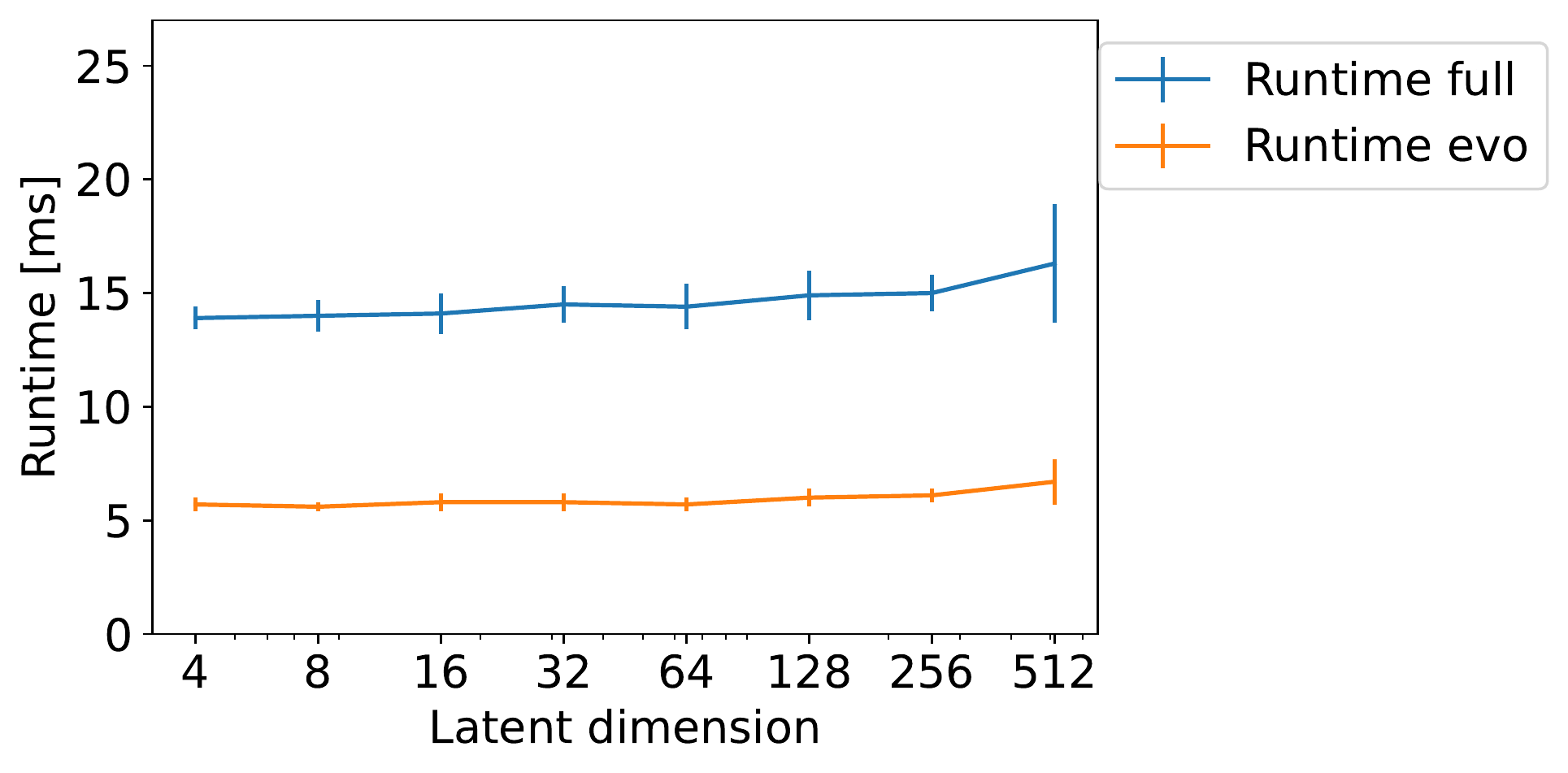}
    \caption{Runtime vs. latent dimension in 1D scenario}
    \label{fig:ablation_1D_runtime}
\end{subfigure}
\begin{subfigure}{0.44\textwidth}
    \includegraphics[width=\textwidth]{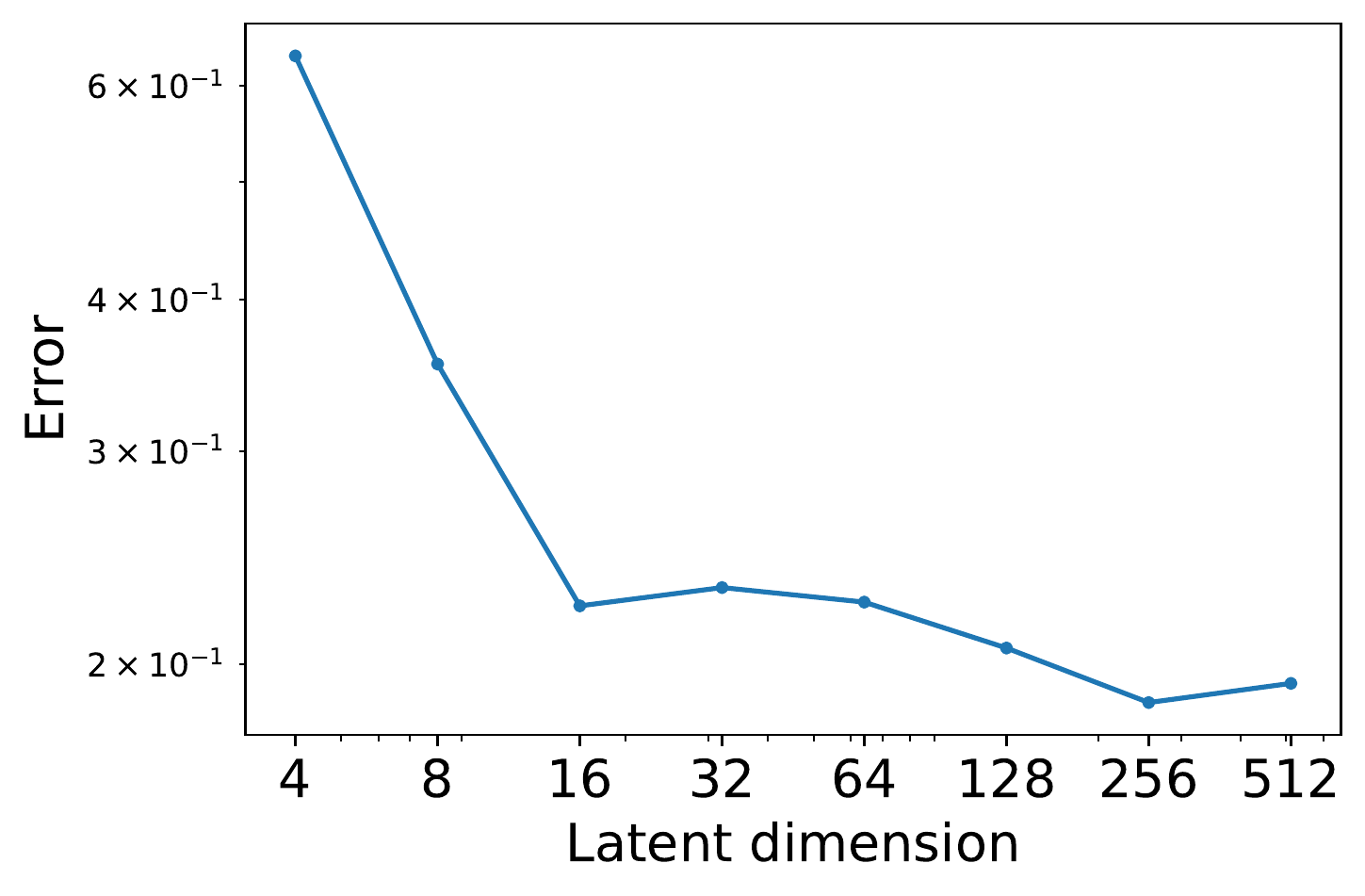}
    \caption{Error vs. latent dimension  in 2D scenario.}
    \label{fig:ablation_2D_error}
\end{subfigure}
\begin{subfigure}{0.55\textwidth}
    \includegraphics[width=\textwidth]{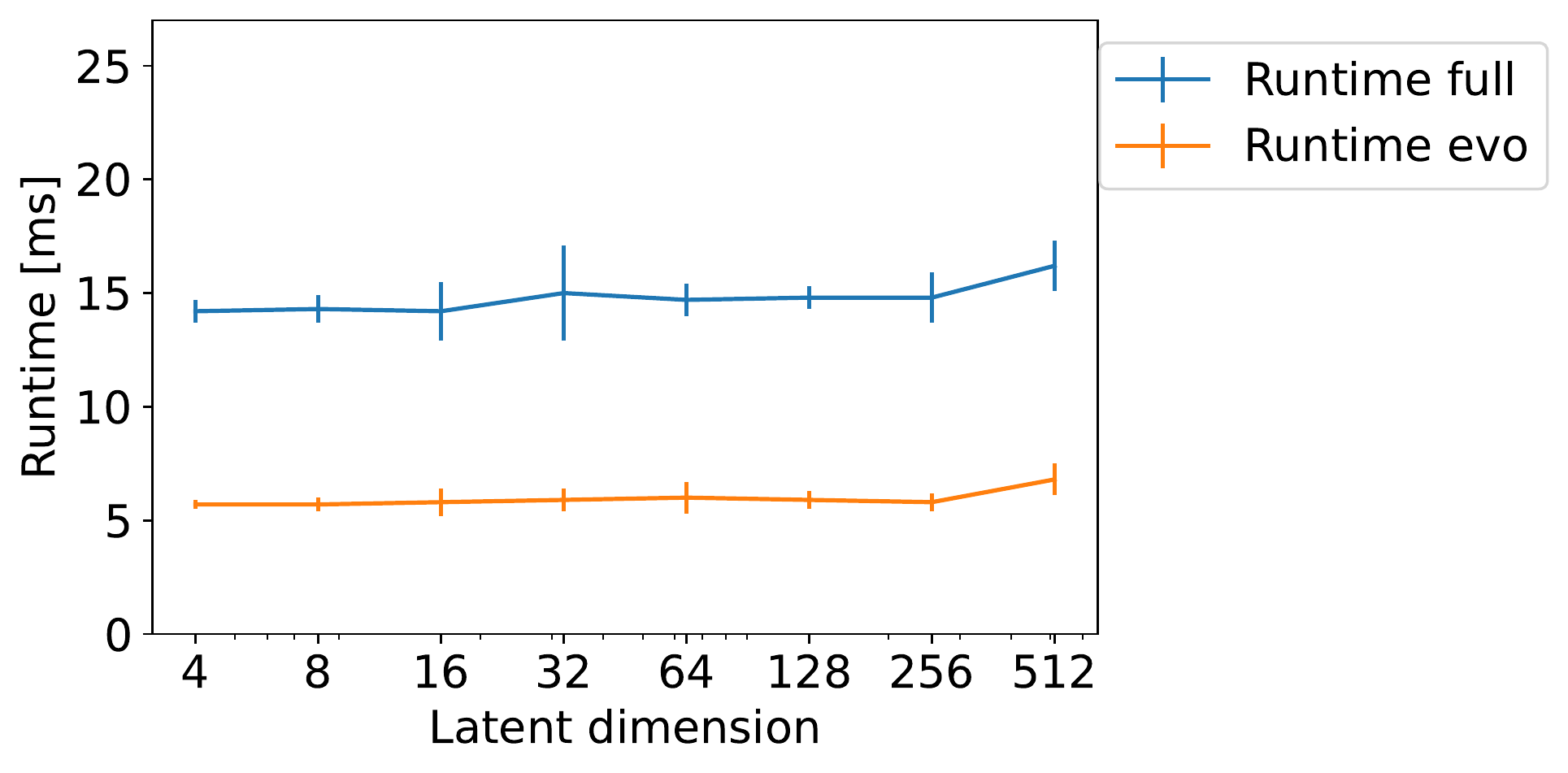}
    \caption{Runtime vs. latent dimension in 2D scenario}
    \label{fig:ablation_2D_runtime}
\end{subfigure}
\caption{Error vs. latent dimension $d_z$ for (a) 1D and (c) 2D scenario, and runtime vs. latent dimension $d_z$ for (b) 1D and (d) 2D scenario. We see that in 1D, the Error stays near optimum with latent size in [16, 128], and goes up outside the range. The runtime evo have a slight decreasing trend from latent dimension at 512 to 256, and stays relatively flat. For 2D, the Error decreases with increasing latent dimension, reaching an optimum at $d_z=256$, and then slightly increases. Its runtime full have a slight decrease from latent dimension of 512 down to 256, and otherwise stays relatively flat.}
\label{fig:ablation}
\end{figure}

\xhdr{More details in the ablation study experiments in Sec. \ref{sec:ablation}} For the ablation ``Pretrain with $L_\text{recons}$'', we pretrain the encoder and decoder with $L_\text{recons}$ for certain number of epochs, then freeze the encoder and decoder and train the latent evolution model and static encoder with $L_\text{consistency}$. Here the $L_\text{multi-step}$ is not valid since the encoder and decoder are already trained and frozen. For both 1D and 2D, we search hyperparameters of pretraining with $\{25,50,100\}$, and choose the model with the best validation performance.

\section{Broader social impact}
\label{app:social_impact}
Here we discuss the broader social impact of our work, including its potential positive and negative aspects, as recommended by the checklist. On the positive side, our work have huge potential implication in science and engineering, since many important problems in these domains are expressed as temporal PDEs, as discussed in the Introduction (Sec. \ref{sec:introduction}). Although this work focus on evaluating our model in standard benchmarks, the experiments in Appendix \ref{app:3d_nv} also show the scalability of our method to problems with millions of cells per time steps under turbulent dynamics. Our \proj can be applied to accelerate the simulation and inverse optimization of the PDEs in science and engineering, \textit{e.g.} weather forecasting, laser-plasma interaction, airplane design, etc., and may significantly accelerate such tasks.

We see no obvious negative social impact of our work. As long as it is applied to the science and engineering that is largely beneficial to society, our work will have beneficial effect.

\section{Pareto efficiency of FNO vs. \proj}
\label{app:pareto}
The following Table \ref{tab:g1} shows the comparison of performance of FNO with varying hyperparameters. The hyperparameter search is performed on a 1D representative dataset E2-50. We evaluate the models (with varying hyperparameters) using the metric of the cumulative error and runtime. The most important hyperparameters for FNO are the ``modes'', which denotes the number of Fourier frequency modes, and ``width'', which denotes the channel size for the convolution layer in the FNO.

\renewcommand{\thetable}{S\arabic{table}}
\makeatother
\setcounter{table}{7}
\begin{table}[h]
\centering
\caption{Performance evaluation with FNO hyperparameter search on 1D dataset (E2-50 scenario.)}
\begin{tabular}{l|c|c|c}
\hline
\makecell{\textbf{FNO setting}}                         & \makecell{\textbf{cumulative}\\ \textbf{error}} & \makecell{\textbf{runtime} \\ \textbf{(full) (ms)}} & \makecell{\textbf{\# parameters}}  \\
\hline
\makecell{modes=16, width=64 \\ (default setting)} & 2.379            & 21.2$\pm$ 6.9         & 292249         \\
\hline
modes=16, width=128                  & 3.107            & 21.7$\pm$ 4.3         & 1138201        \\
modes=16, width=32                   & 2.695            & 22.1$\pm$ 7.4         & 78169          \\
modes=16, width=16                   & 2.755            & 21.0$\pm$ 5.7         & 23353          \\
modes=16, width=8                    & 4.992            & 17.9$\pm$ 1.2         & 9001           \\
modes=20, width=128                  & 2.804            & 20.9$\pm$ 1.1         & 1400345        \\
modes=20, width=64                   & 2.626            & 19.3$\pm$ 0.9         & 357785         \\
modes=12, width=64                   & 2.899            & 19.6$\pm$ 2.2         & 226713         \\
modes=8, width=64                    & 2.240            & 19.7$\pm$ 1.3         & 161177         \\
modes=4, width=64                    & 2.326            & 19.2$\pm$ 0.9         & 95641          \\
modes=8, width=32                    & 2.366            & 18.2$\pm$ 1.0         & 45401          \\
modes=8, width=16                    & 2.505            & 18.1$\pm$ 1.2         & 15161          \\
modes=8, width=8                     & 5.817            & 18.4$\pm$ 1.2         & 6953          \\
\hline
\end{tabular}
\label{tab:g1}
\end{table}

We also perform hyperparameter search on a 2D representative dataset with $\nu=10^{-5}$. Table \ref{tab:g2} shows the comparison of performance of FNO with varying hyperparameters. Hyperparameters to be varied and metrics for the evaluation are same as that of Table \ref{tab:g1}.
\begin{table}[h]
\centering
\caption{Performance evaluation with FNO hyperparameter search on 2D dataset ($\nu=10^{-5}$ scenario.)}
\begin{tabular}{l|c|c|c}
\hline
\makecell{\textbf{FNO setting }}                         & \makecell{\textbf{L2 error}} & \makecell{\textbf{runtime }\\\textbf{(full)(ms)}} & \makecell{\textbf{\# parameters}} \\
\hline
\makecell{modes=12, width=20 \\ (default setting)} & 0.1745              & 42.7 $\pm$ 10.9        & 465717        \\
\hline
modes=12, width=40                   & 0.1454              & 42.7 $\pm$ 4.2         & 1855977       \\
modes=12, width=10                   & 0.2016              & 40.3 $\pm$ 5.4         & 117387        \\
modes=12, width=5                    & 0.2398              & 45.5 $\pm$ 7.4         & 29922         \\
modes=16, width=20                   & 0.1710              & 43.7 $\pm$ 4.2         & 824117        \\
modes=8, width=20                    & 0.1770              & 43.1 $\pm$ 3.1         & 209717        \\
modes=4, width=20                    & 0.1997              & 43.2 $\pm$ 4.8         & 56117         \\
modes=8, width=10                    & 0.2109              & 42.2 $\pm$ 4.8         & 53387         \\
modes=8, width=5                     & 0.2415              & 43.3 $\pm$ 4.3         & 13922        \\
\hline
\end{tabular}
\label{tab:g2}
\end{table}

\renewcommand{\thefigure}{S\arabic{figure}}
\makeatother
\setcounter{figure}{7}
\begin{figure}[t]
\centering
\begin{subfigure}{0.49\textwidth}
    \includegraphics[width=\textwidth]{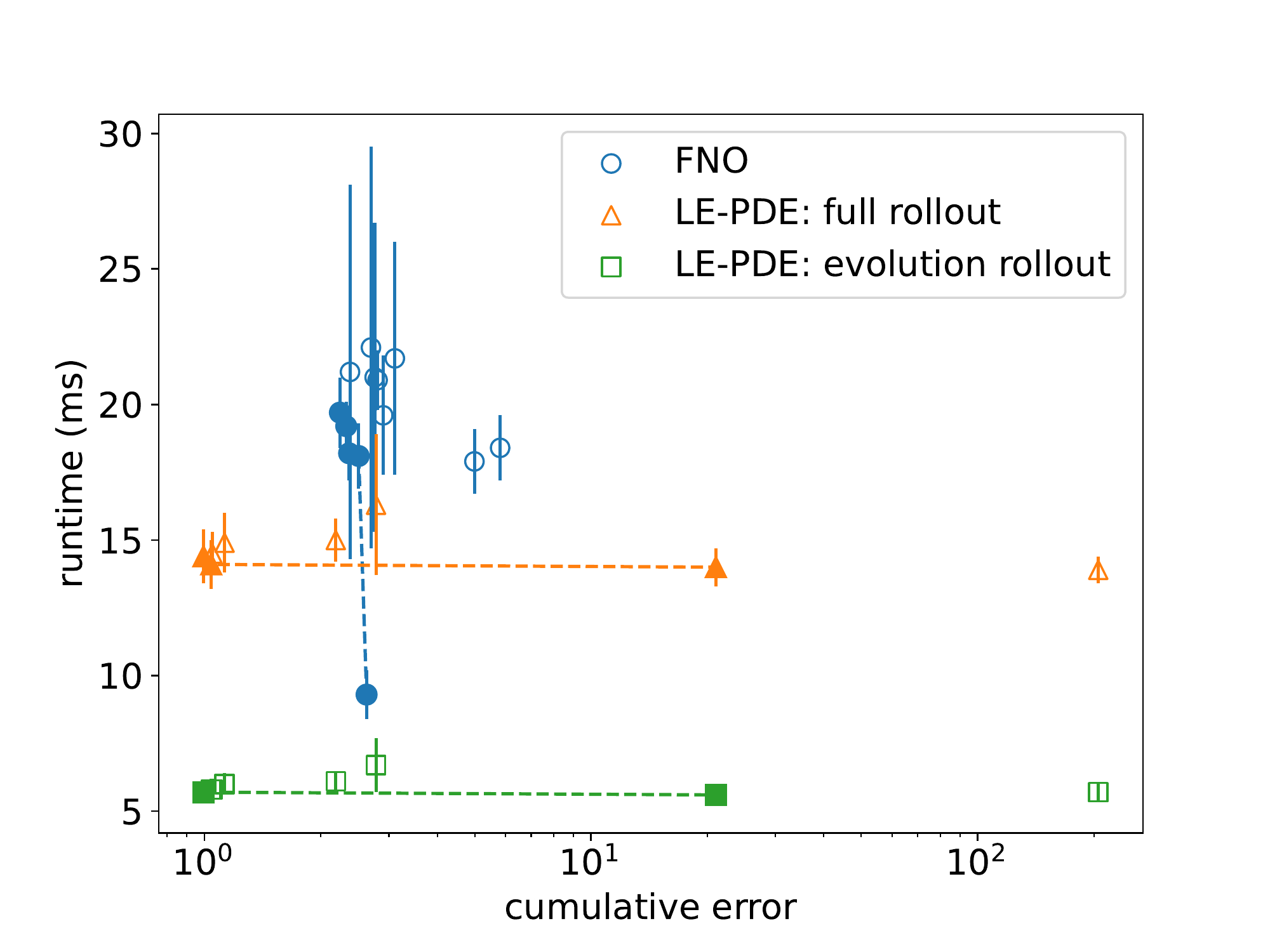}
    \caption{1D dataset.}
    \label{fig:1d_err_run}
\end{subfigure}
\begin{subfigure}{0.49\textwidth}
    \includegraphics[width=\textwidth]{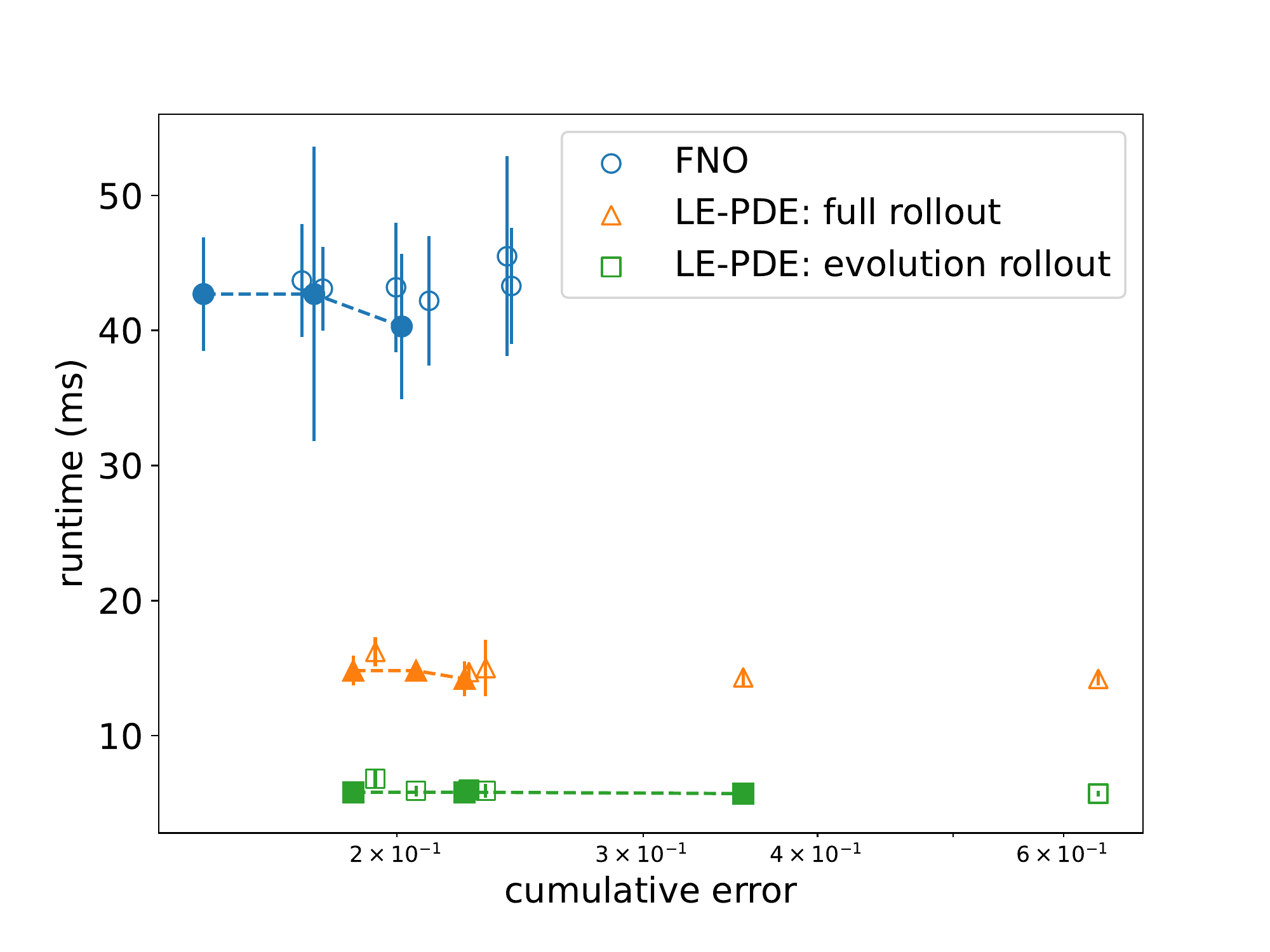}
    \caption{2D dataset.}
    \label{fig:2d_err_run}
\end{subfigure}
\caption{Comparison of trade-off between cumulative error and runtime of \proj and FNO for 1D and 2D dataset. Dotted line connected to filled marker is Pareto frontier for respective model.}
\label{fig:compare_1d_tradeoff}
\end{figure}

\begin{figure}[t]
\centering

\begin{subfigure}{0.49\textwidth}
    \includegraphics[width=\textwidth]{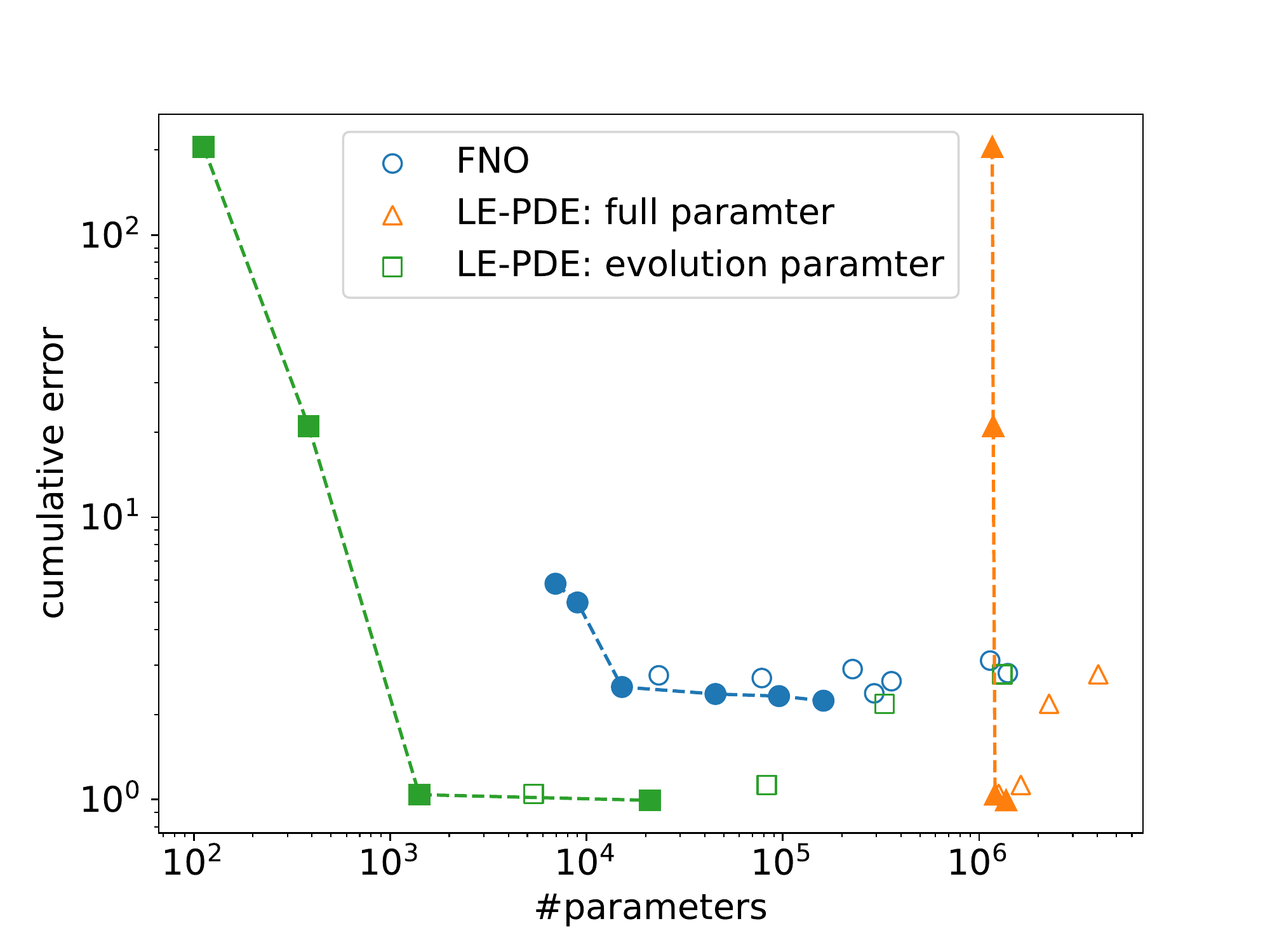}
    \caption{1D dataset.}
    \label{fig:1d_para_err}
\end{subfigure}
\begin{subfigure}{0.49\textwidth}
    \includegraphics[width=\textwidth]{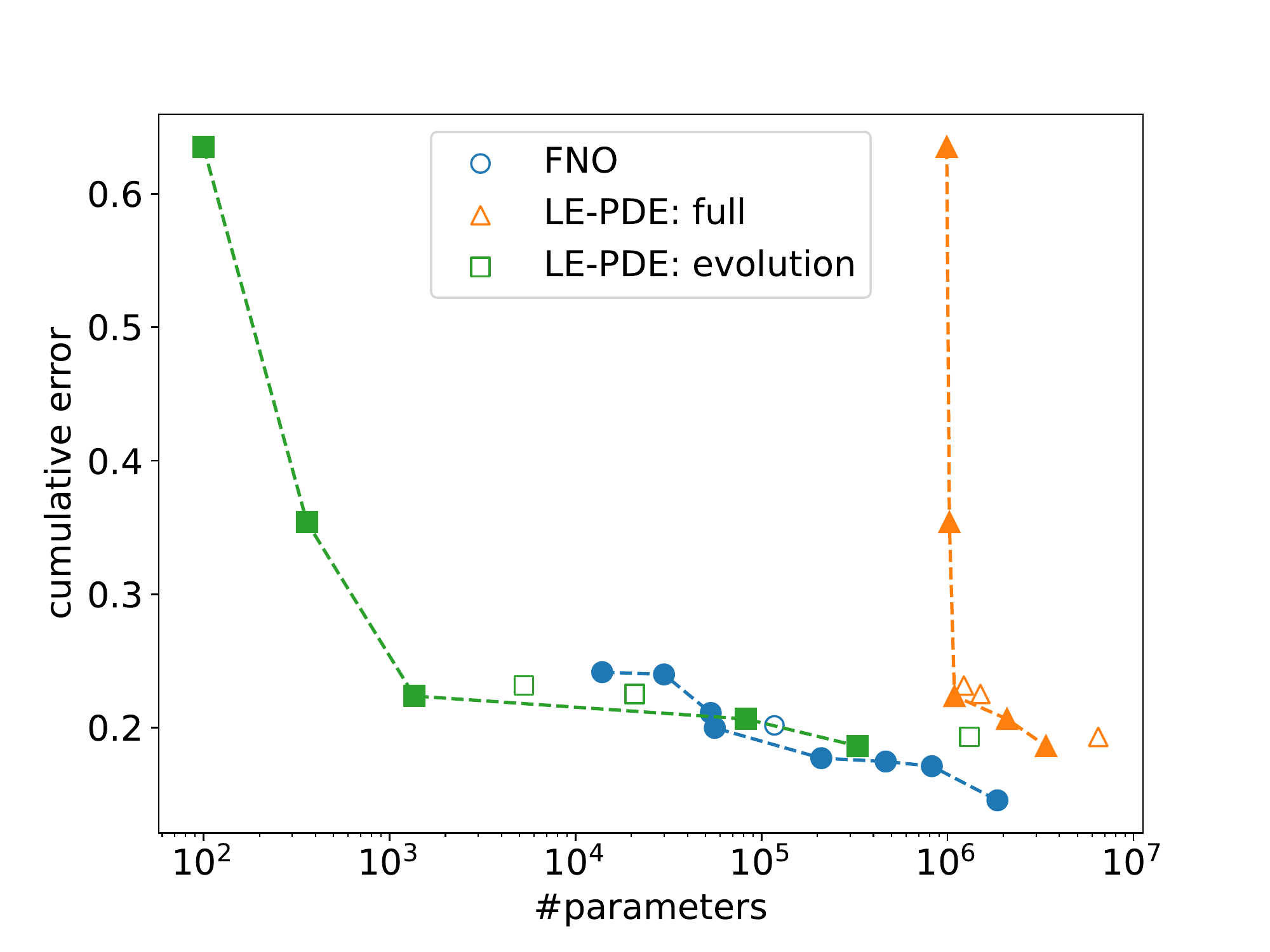}
    \caption{2D dataset.}
    \label{fig:2d_para_err}
\end{subfigure}
\caption{Comparison of trade-off between number of parameters and cumulative error of \proj and FNO for 1D and 2D dataset. Dotted line connected to filled marker is Pareto frontier for respective model.}
\label{fig:compare_2d_tradeoff}
\end{figure}

We can compare the above two tables with Table \ref{table:ablation_latent_size_1d} and \ref{table:ablation_latent_size_2d}. We also create plots Figure \ref{fig:compare_1d_tradeoff} and \ref{fig:compare_2d_tradeoff} that compare the trade-off between several metrics shown in the tables for \proj and FNO. Note that we provide both the total number of parameters (second last column) and number of parameters for latent evolution model (last column). The latter is also a good indicator since during long-term evolution, the latent evolution model is autoregressively applied while the encoder and decoder are only applied once. So the latent evolution model is the deciding component of the long-term evolution accuracy and runtime. 

From the comparison, we see that:
\begin{itemize}
\item For 1D dataset, \proj Pareto-dominates FNO in error vs. runtime plot (Fig. \ref{fig:compare_1d_tradeoff}(a)). FNO’s best cumulative error is 2.240, and runtime is above 17.9ms, over the full hyperparameters combinations (number of parameter varying from 6953 to 1.4M). In comparison, our \proj achieves much better error and runtime over a wide parameter range: for $d_z$ from 16 to 64, LE-PDE's cumulative error $\leq$ 1.05, runtime $\leq$ 14.5ms, latent runtime $\leq$ 5.8ms, (which uses 1408 to 82944 number of parameters for latent evolution model, and 1.2-1.4M total parameters). In terms of cumulative error vs. \#parameter plot (Fig. \ref{fig:compare_2d_tradeoff}(a)), the LE-PDE with evolution model typically has less parameters than FNO, which in turn also have less parameters than LE-PDE with full model. This makes sense, as the latent evolution requires much less parameters. Adding the encoder and decoder, \proj may have more  \#parameters. But still it is the evolution parameter that is the most important for long-term evolution.

\item For 2D dataset, FNO’s cumulative error is slightly better than \proj, but its runtime is significantly larger (Fig. \ref{fig:compare_1d_tradeoff}(b)). Concretely, the best FNO achieves an error of 0.1454 while the best LE-PDE’s error is 0.1861. FNO’s runtime is above 40ms, while LE-PDE’s runtime is generally below 15ms and latent evolution runtime is below 6ms. \proj uses larger total number of parameters but much less number of parameters for latent evolution model. Also, similar to 1D, in terms of  error vs. \#parameter plot (Fig. \ref{fig:compare_2d_tradeoff}(b)), the LE-PDE with evolution model typically has much less parameters than FNO, which in turn also typically have less parameters than LE-PDE with full model.
\end{itemize}

\xhdr{Which family of PDEs can our \proj apply:} we can think of a PDE as a ground-truth model that evolves the \emph{state} of a physical system. Typically, the states show more global, dominant features, and can be described by a state vector with much smaller dimension than the original discretization. Our LE-PDE exploit this \emph{compressivity of state} to evolve the system in latent space and achieve speedup, and as long as the PDE does not significantly increase the spatial complexity of the state as it evolves (e.g. developing finer and finer spatial details as in 2-stream instability in of plasma \cite{roberts1967nonlinear}), our method can apply. Most of the PDEs satisfy the above requirements that the state are compressible and does not significantly increase its complexity, so our LE-PDE can be apply to most PDEs. Since any compression of state can incur a possible increase of error (possibly large or small, as the Pareto frontier of "error vs. runtime" and "error vs. \#parameter" in Fig. \ref{fig:compare_1d_tradeoff} and \ref{fig:compare_2d_tradeoff} show for our LE-PDE and FNO), the more important/relevant question is then "what is the tradeoff of error vs. runtime we want for the given PDE", since we can design the encoder of LE-PDE with varying amount of compression. For example, we can design an encoder with minimal compression, so runtime reduction is low but can guarantee to retain low error, or with a much more aggressive compression (like in our 2D and 3D experiments), but can still achieve minimal increase of error. The amount of compression is a hyperparameter which can be obtained via validation set. Theoretically studying the best amount of compression that achieves a good tradeoff will be left for an exciting future work.

\section{Comparison of LE-PDE with LFM}
\label{app:lfm}
To compare our \proj with the Latent Field Model method (LFM) proposed in \cite{sanchez2020learning2}, we perform additional experiments in the representative 1D   and 2D datasets in Section \ref{sec:ablation}. We perform the ablation study where we (a) remove MLP in our model, (b) use LFM objective but maintain MLP, and (c) full LFM: remove MLP, use LFM objective, while all other aspects of training is kept the same. We use PyTorch’s jvp function in autograd to compute the Jacobian-vector product and carefully make sure that our implementation is correct. Table \ref{tab:g5} is the comparison table.
\begin{table}[h]
\centering
\caption{Performance comparison of \proj with LFM, for 1D dataset E2-50 scenario.}
\resizebox{\textwidth}{!}{\begin{tabular}{l|c|c|c|c|c}
\hline
\makecell{\textbf{\proj}\\ \textbf{setting}}                                & \makecell{\textbf{cumulative}\\ \textbf{error}} & \makecell{\textbf{runtime}\\ \textbf{(full) (ms) }}                & \makecell{\textbf{runtime}\\ \textbf{(evolution)}\\ \textbf{(ms)}}           & \makecell{\textbf{\# parameters}}                   & \makecell{\textbf{\# parameters for}\\ \textbf{latent evolution}\\ \textbf{model}} \\
\hline
LE-PDE (ours)                                 & 1.127            &  14.9  $\pm$ 1.1 &  6.0  $\pm$ 0.4 &  1630976 &  82944            \\
\hline
(a) without MLP                               & 7.930            & 17.2  $\pm$ 6.0                         &  8.3  $\pm$ 0.4 &  2730368 &  1580544          \\
(b) with LFM objective                        & 58.85            & 15.7  $\pm$ 1.5                         &  6.5  $\pm$ 0.6 &  1630976 &  82944            \\
\makecell{(c) full LFM: without MLP, \\ with LFM objective} & 26.12            & 15.7  $\pm$ 1.3                         &  8.4  $\pm$ 0.7 &  2730368 &  1580544\\
\hline
\end{tabular}
}
\label{tab:g5}
\end{table}

Table \ref{tab:g6} shows the comparison result of \proj with LFM obtained by performing additional experiments on the representative 2D dataset in Section \ref{sec:ablation}.

\begin{table}[h]
\centering
\caption{Performance comparison of \proj with LFM, for 2D dataset $\nu=1e$-$5$ scenario.}
\resizebox{\textwidth}{!}{\begin{tabular}{l|c|c|c|c|c}
\hline
\makecell{\textbf{\proj setting}}                                & \makecell{\textbf{cumulative}\\ \textbf{error}} & \makecell{\textbf{runtime}\\ \textbf{(full) (ms)}}                 & \makecell{\textbf{runtime}\\ \textbf{(evolution)}\\ \textbf{(ms)}}           & \makecell{\textbf{\# parameters }}                  & \makecell{\textbf{\# parameters for}\\ \textbf{evolution model}} \\
\hline
\proj (ours)                                 & 0.1861           &  14.8  $\pm$ 1.1 &  5.8  $\pm$ 0.4 &  3384944 &  328960           \\
\hline
(a) without MLP                               & 0.2120           &  16.6  $\pm$ 2.2 &  9.2  $\pm$ 0.8 &  2126960 &  1181184          \\
(b) with LFM objective                        & 0.4530           &  15.8  $\pm$ 2.3 &  6.2  $\pm$ 0.6 &  3384944 &  328960           \\
\makecell{(c) full LFM: without MLP, \\with LFM objective} & 0.6315           &  16.2  $\pm$ 1.9 &  9.1  $\pm$ 0.4 &  2126960 &  1181184     \\
\hline
\end{tabular}
}
\label{tab:g6}
\end{table}

From the above tables, we see that without MLP, it actually results in worse performance (ablation (a)), and with LFM objective, the error is larger, likely due to that the dataset are quite chaotic and LFM may not adapt to the large time range in these datasets.

\section{Influence of varying noise amplitude}
\label{sec:abnoise1d}
Here, we perform additional experiments on how the noise affects the performance, on the representative 1D used in Section \ref{sec:ablation} “Ablation Study”. Table \ref{tab:g8} shows the results. Specifically, we add random fixed Gaussian noise to the training, validation and test sets of the dataset, with varying amplitude. The noise is independently added to each feature of the dataset. It is also “fixed” in the sense that once added to the dataset, the noise is freezed and not re-sampled. This mimics the real world setting where random observation noise can corrupt the observation and we never have the ground-truth data to train and evaluate from.
\begin{table}[h]
\centering
\caption{Evaluation of cumulative error of \proj on 1D dataset (E2-50 scenario) with varying noise amplitude. The amplitude is the standard deviation of the diagonal Gaussian and the value range of the state $u(t,x)$ is within $[-2,2]$.}
\begin{tabular}{c|c}
\hline
\makecell{\textbf{Noise amplitude}} & \makecell{\textbf{cumulative error}} \\
\hline
$0$ (default)     & $1.127$            \\
$10^{-5}$         & $1.253$            \\
$10^{-4}$         & $1.268$            \\
$10^{-3}$         & $1.456$            \\
$10^{-2}$         & $2.612$            \\
$2\times10^{-2}$         & $4.102$            \\
$5\times10^{-2}$          & $9.228$           \\
\hline
\end{tabular}
\label{tab:g8}
\end{table}

We also perform experiments similar to Section \ref{sec:abnoise1d} on a 2D representative dataset used in Section \ref{sec:ablation} “Ablation Study”. Table \ref{tab:g9} shows the results.
\begin{table}[h]
\centering
\caption{Evaluation of cumulative error of \proj on 2D dataset ($\nu=10^{-5}$ scenario) with varying noise amplitude. The amplitude is the standard deviation of the diagonal Gaussian and the value range of the state $u(t,x)$ is within $[-2,2]$.}
\begin{tabular}{c|c}
\hline
\makecell{\textbf{Noise amplitude}} & \makecell{\textbf{cumulative error}} \\
\hline
$0$ (default)     & $0.1861$           \\
$10^{-5}$             & $0.1880$           \\
$10^{-4}$            & $0.1862$           \\
$10^{-3}$            & $0.1866$           \\
$10^{-2}$             & $0.1897$           \\
$2\times10^{-2}$              & $0.1875$          \\
$5\times10^{-2}$            & $0.1910$           \\
$10^{-1}$             & $0.2012$         \\
\hline
\end{tabular}
\label{tab:g9}
\end{table}

Note that the value range of both datasets are within $[-2,2]$. From Table \ref{tab:g8}, we see that LE-PDE’s cumulative error stays excellent ($\le 1.456$) with noise amplitude $\le 10^{-3}$, much smaller than state-of-the-art MP-PDE’s error of $1.63$ and FNO-PF’s $2.27$. Even with noise amplitude of $10^{-2}$, the LE-PDE’s error of $2.612$ still remains reasonable.

From Table \ref{tab:g9}, we see that LE-PDE is quite resilient to noise, with error barely increases for noise amplitude up to $2\times10^{-2}$, and only shows minimal increase at noise level of $10^{-1}$. As a context, U-Net’s error is $0.1982$ and TF-Net’s error is $0.2268$ (Table $2$ in main text).

In summary, in the 1D and 2D datasets, we see that LE-PDE shows good robustness to Gaussian noise, where the performance is reasonable where the ratio of noise amplitude to the value range can go up to 0.25\% in 1D and 2.5\% in 2D. The smaller robustness in the 1D Burgers' dataset may be due to that it is a 200-step rollout and the noise may make the model uncertain about the onset of shock formation.

\section{Ablation of LE-PDE using pretrained autoencoder or VAE}
\label{app:encoder}
In addition, we perform two ablation experiments that explore performing data reduction first and then learn the evolution in latent space: (a) pretrain an autoencoder with states from all time steps, then freeze the autoencoder and train the latent evolution model. This mimics the method in [79]. (b) the encoder and decoder of LE-PDE is replaced with a VAE, first pre-trained with ELBO on all time steps, then freeze the encoder and decoder and train the latent evolution model. All other aspects of the model architecture and training remains the same. The result is shown in the Table \ref{tab:g10} and Table \ref{tab:g11} for the 1D and 2D datasets in Section \ref{sec:ablation} of “Ablation study”.
\begin{table}[h]
\centering
\caption{Ablation study of \proj using pretrained autoencoder or VAE, for 1D dataset (E2-50 scenario.)}
\begin{tabular}{c|c}
\hline
\makecell{\textbf{\proj setting}}           & \makecell{\textbf{Cumulative error}} \\
\hline
\proj (ours)            & 1.127            \\
\hline
(a) pretrain autoencoder & 1.952            \\
(b) pretrained VAE       & 1.980           \\
\hline
\end{tabular}
\label{tab:g10}
\end{table}

\begin{table}[h]
\centering
\caption{Ablation study of \proj using pretrained autoencoder or VAE, for 2D dataset ($\nu=10^{-5}$ scenario.)}
\begin{tabular}{c|c}
\hline
\makecell{\textbf{\proj setting }}          & \makecell{\textbf{Cumulative error}} \\
\hline
\proj (ours)            & 0.1861           \\
\hline
(a) pretrain autoencoder & 0.2105           \\
(b) pretrained VAE       & 0.2329           \\
\hline
\end{tabular}
\label{tab:g11}
\end{table}

From Table \ref{tab:g10} and \ref{tab:g11}, we see that performing pre-training results in a much worse performance, since the data reduction only focuses on reconstruction, without consideration for \emph{which} latent state is best used for evolving long-term into the future. On the other hand, our LE-PDE trains the components jointly with a novel objective that not only encourages better reconstruction, but also long-term evolution accuracy both in latent and input space. We also see that VAE as data-reduction performs worse than autoencoder, since the dynamics of the system is deterministic, and having a stochasticity from the VAE does not help.

\section*{References}
[74] Y. Wu and K. He, “Group normalization,” in \textit{Proceedings of the European conference on computer vision (ECCV)}, 2018, pp. 3–19. 

[75] D.-A. Clevert, T. Unterthiner, and S. Hochreiter, “Fast and accurate deep network learning by exponential linear units (elus),” \textit{International Conference on Learning Representations}, 2016, [Online]. Available: https://arxiv.org/abs/1511.07289. 

[76] L. Jing, J. Zbontar \textit{et al.}., “Implicit rank-minimizing autoencoder,” in \textit{Advances in Neural Information Processing Systems}, vol. 33, pp. 14736–14746, 2020. 

[77] A. A. Kaptanoglu, K. D. Morgan, C. J. Hansen, and S. L. Brunton, “Physics-constrained, low dimensional models for magnetohydrodynamics: First-principles and data-driven approaches,” in \textit{Physical Review E}, vol. 104, no. 1, p. 015206, 2021. 

[78] S. Yang, X. He, and B. Zhu, “Learning physical constraints with neural projections,” in \textit{Advances in Neural Information Processing Systems}, vol. 33, pp. 5178–5189, 2020. 

[79] I. Loshchilov and F. Hutter, “SGDR: Stochastic gradient descent with warm restarts,” in \textit{International Conference on Learning Representations (Poster)}, 2017, [Online]. Available: https://arxiv.org/abs/1608.03983.

\end{document}